\documentclass[final,3p,times]{elsarticle}
\usepackage{amsmath}
\usepackage{upgreek}
\usepackage{setspace}
\usepackage{nomencl}
\usepackage[utf8]{inputenc}
\usepackage[T1]{fontenc}
\usepackage{subcaption}
\usepackage{amssymb}
\usepackage{natbib}
\usepackage{xcolor}
\usepackage{graphicx}
\usepackage[black]{changes}
\journal{Mechanism and Machine Theory}
\begin{document}
\begin{frontmatter}
\title{A note on synthesizing geodesic based contact curves}
\author{Rajesh Kumar, Sudipto Mukherjee}

\address{Department of Mechanical Engineering, Indian Institute of Technology Delhi, India}

\begin{abstract}
The paper focuses on synthesizing optimal contact curves that can be used to ensure a rolling constraint between two bodies in relative motion. We show that geodesic based contact curves generated on both the contacting surfaces are sufficient conditions to ensure rolling. The differential geodesic equations, when modified, can ensure proper disturbance rejection in case the system of interacting bodies is perturbed from the desired curve. A corollary states that geodesic curves are generated on the surface if rolling constraints are satisfied. Simulations in the context of in-hand manipulations of the objects are used as examples. 
\end{abstract}

\begin{keyword}
Geodesic Curves \sep Contact Curves \sep Rolling
\end{keyword}

\end{frontmatter}

\section{Introduction}
Robotic grasping and manipulation is achieved through the interaction of robotic fingers with objects \citep{andrychowicz2018learning, sundaralingam2018geometric, chong1993generalized}. In order to carry out the manipulation, the robotic fingers either roll \cite{paljug1994control, bicchi1995dexterous, maekawa1995tactile}, slide \cite{shi2017dynamic, spiers2018variable} or stick \cite{chavan2018stable} on the object surface or use a combination of all three motions \cite{cherif1999planning}. Rolling contacts during in-hand manipulation are known to accord a larger workspace to the object being manipulated by robotic fingers \cite{weili1994workspace}. The relative motion of the object with respect to the fingertips generates contact curve on the object as well as on the fingertips \cite{montana1988kinematics}. The gross motion of the object is related to the contact curves if the type of relative motion (sliding, rolling, or a combination of sliding and rolling) is known \cite{montana1988kinematics}.
 \par
\begin{figure}[t]
    \centering
    \includegraphics[trim={2cm 17cm 0cm 1cm}, clip, scale=0.7]{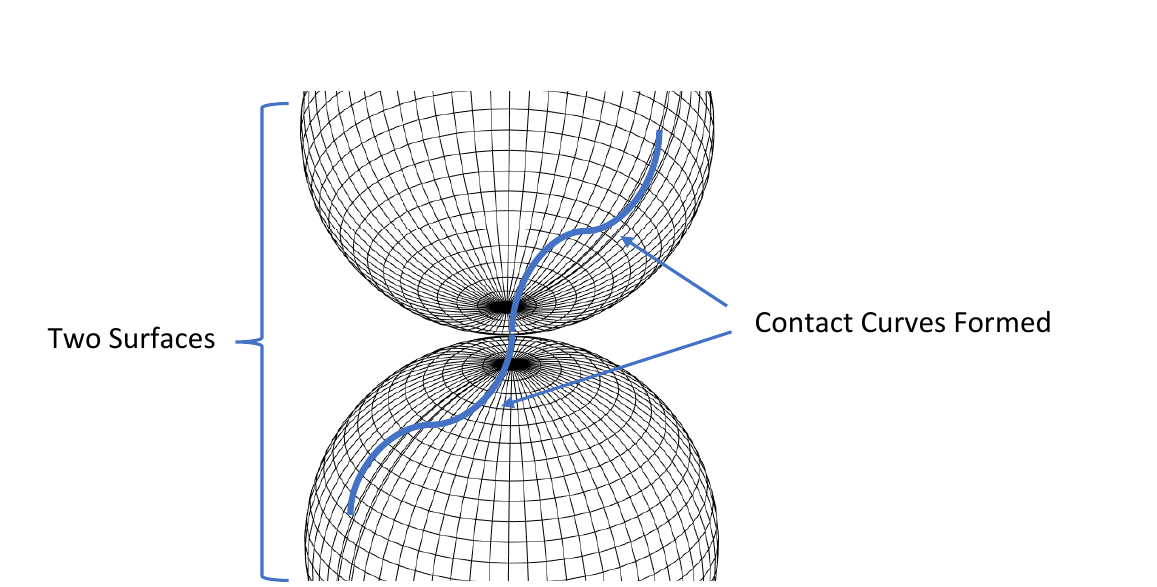}
    \caption{Exemplar contact curves formed on two surfaces moving relatively in contact}
    \label{example}
\end{figure}

The contact curve is the locus of the instantaneous point of contact between two contacting surfaces in relative motion and is formed on both the surfaces (as shown in Figure \ref{example}).
For planar two dimensional objects, the contact curves are restricted to lie along the contact edge only. However, for three-dimensional contacts, as encountered during in-hand manipulation of objects, spatial contact curves are formed. There have been an interest in analysing the contact curves generated for a specific type of relative motion. The differential constraint equations constraining the relative motion of objects leading to the generation of contact curves were formulated for both rolling and sliding during the late 1980s \cite{montana1988kinematics}. These algebraic equations relate the time derivative of the coordinates attached to the surfaces to the instantaneous linear and angular velocities of the two rigid bodies. Montana's contact curve equations \cite{montana1988kinematics} have also been utilised to formulate control strategies to ensure that an enhanced grasp pose is achieved \cite{han1997dextrous}. The grasp angle is defined as the angle between the contact force vector and surface normal at the point of contact. The control algorithm \cite{han1997dextrous} was based on minimizing the grasp angle during in-hand manipulation.  Although the generalised contact equations have not seen much implementation, the rolling contact equations have been utilised for simulation and control of robotic fingers \cite{bicchi1995dexterous} in contact with the object. Choudhary and Lynch \cite{choudhury2002rolling} have also used these equations to synthesize $\textit{single input}$ control of a sphere in a bowl. The equations were also utilised to plan contact curves on the surface of the sphere~\cite{svinin2008motion}. Although Montana's equations \cite{montana1988kinematics} have been widely used for simulation and control purposes, they relate only the first-order derivatives of the contact curves to the velocities of the object. Sankar et al. \cite{sankar1996velocity} associated the second-order derivatives of the surface variables (like the curvature form) with the acceleration of the body. This conjecture was useful to associate surface geometry with the dynamics of the system. \par Contact curves have also been utilised to formulate exact forward and inverse kinematics during rolling motion \cite{cui2017hand}, and to formulate a polynomial order computation of motion under rolling contact leading to reduction in computation time \cite{cui2015polynomial}. The contact acceleration equations derived in \cite{sankar1996velocity} have been used to formulate feedback linearizable control structures that admit control of the gross motion of the object as well as the differential motion of the fingers on the object \cite{sarkar1997dynamic}. The method was presented by Sarkar et al. \cite{sarkar1997dynamic}, where the motion of the fingers was used to control the net object motion and the contact curve generated on the object surface.   

\par Researchers have also focussed on analysing the nature of the contact curves formed for a particular type of relative motion. An adjoint-based approach was developed for a sliding - rolling system \cite{cui2015sliding}. A Darboux-frame based approach has been formulated for rolling contact motion \cite{cui2010darboux}.  Contact curves have also been utilised to plan the initial motion of sliding/ rolling systems to avoid obstacles \cite{jia2016planning} while rolling. \par Though the use of contact curves have been demonstrated, the contact curves have not been utilised to get a desired relative motion (like rolling). The methods stated above do not focus on generating contact curves to get the desired motion. However, there have been few instances where the contact curves were related to the type of motion. During the 1990s, a finger gaiting methodology formulated by Han et al. \cite{han1998dextrous} concluded that rolling on great circles on a spherical object will lead to $\textit{great circle}$ contact curves on the spherical finger as well. This was one of the first instances where the generation of a particular type of contact curve was related to the rolling motion. Later, Lei et al. \cite{cui2009coordinate} utilised contact curves to show that if two bodies are in rolling constraint, then two geometrical invariant properties (torsion form and the curvature form) completely define the relative angular velocity. It was shown that the difference between the geodesic curvatures of the two contact curves formed on contacting bodies is zero during rolling.
\par A $\textit{Geodesic curve}$ is formally defined as the path with the shortest distance between two points on a surface \cite{kumar2003geodesic}, and have been studied intensely in the field of differential geometry. The same curve can also be defined as the trace of a particle moving on a surface with acceleration purely normal to the surface. In an extension to dimensions higher than $\mathbb{R}^3$, geodesic contact curves are defined in terms of the second-order Christoffel symbols \cite{haw1983geodesic} as per equation \ref{geodesicdef}.
\begin{equation}
    \label{geodesicdef}
    \frac{d^2\psi_{i}}{ds^2} + \Gamma^i_{jk}\frac{d\psi_j}{ds}\frac{d\psi_k}{ds} = 0
\end{equation}
 where $\psi_t$ represents the $t^{th}$ coordinate on the manifold (n-dime nsional) immersed in $\mathbb{R}^m$ for t = 1:n and $\Gamma^i_{jk}$ represents the Christoffel symbols of the second kind for the manifold. Any object surface in three dimensions can be considered as a 2-dimensional manifold immersed in $\mathbb{R}^3$. Hence, two geodesic equations are generated.  For a real object surface (2-D), the symbol $\psi_t$ represents the surface coordinates (called chart variables). Points on the surface can be represented by a 2-D parametric function (chart function). It should be understood that the geodesic equations do not define exact coordinates of contact curves but relate the second-order derivative of the growth of surface coordinates (chart variables) to the growth of the first-order derivative of the coordinates at a given location. In this paper, we prove some results related to rolling motion, focussing on cases where geodesic curves are formed due to typical motion resulting from an object  being manipulated by fingers. This includes proof of guaranteed no-slip motion if the finger and the object contacts move on geodesic curves at all times. Geodesics are not natural attractors and hence do not ensure proper disturbance rejection. A modification to the classical geodesic equation is proposed, which ensures proper disturbance rejection. The modification is done by driving the system (the system which generates the contact curves) to a contracting region \cite{lohmiller1998contraction}. 
\par In this paper, \added{we show that if geodesic based contact curves are generated on the surface of contacting bodies, then rolling constraints are maintained. The classical differential geodesic equations in surface coordinates are modified to synthesize a relationship between the growth of geodesic curves and the relative velocity between the contact frames. A further modification to the classical geodesic equations is shown to reject sliding disturbances between the contact frames.  The methodology provides a kinematic approach to ensure that the relative contact velocity reduces to zero after perturbation. Apart from the proof that geodesic based contact curves result in rolling constraints, a corollary to the proof is also presented. The corollary states that if the contacting bodies demonstrate rolling constraint and one of the contact curve is a geodesic, then the contact curve on the other contacting body will also be a geodesic curve. The derivations of the expressions of relative acceleration and relative velocity (using contact curves) are inspired by the methods in \cite{sankar1996velocity}, but the expressions are reformulated to make them amenable to treatment using classical differential geometry.} \added{The methodology developed can be used to sustain rolling constraints during rolling in-hand manipulation of the object. The fingertips can map the geodesic based trajectories such that the resulting contact curves on the surface of the finger and the surface of the object are geodesics.}\par  \added{Although the method is developed considering inhand manipulation of an object, the method can be used to maintain rolling constraints for any set of contacting bodies and the proofs provided in the paper are applicable to general contacting rigid bodies. The hard fingertip can be replaced by any actuated body (or manipulator) and the interacting object can be fixed or free to be manipulated. A manipulator rolling its end effector over a manifold is one such instance.  The contributions of the paper can be summarized as:}
\begin{itemize}
    \item \added{Understanding the nature of contact between bodies in relative motion when the trace of the contact points on the body surface are geodesics and its correlation to rolling contact. A proof relating the existence of rolling contact between the contacting frames and the synthesis of geodesic based contact curves is presented. The architecture is studied in context of in-hand manipulation of an object. }
    \item \added{ It is shown that when the progress of the contact is guided by a proposed modification to the geodesic based contact equations, it results in rejection of slip disturbances for two bodies in contact. The method can be used for synthesis of practical in-hand manipulation where rolling contact is desired. If the contacting bodies follow the modified geodesic equations, then rolling contact can be regained post slip.}
    \item \added{A proof to a corollary is presented. The corollary states that if geodesic contact curves are known to be maintained on one contacting body, then geodesic based contact curves will be synthesised on the other contacting body as well, if rolling constraints are satisfied at all times.}
\end{itemize}
 \begin{figure}
     \centering
     \includegraphics[width=0.5\textwidth, trim={6cm 20cm 3cm 3cm}]{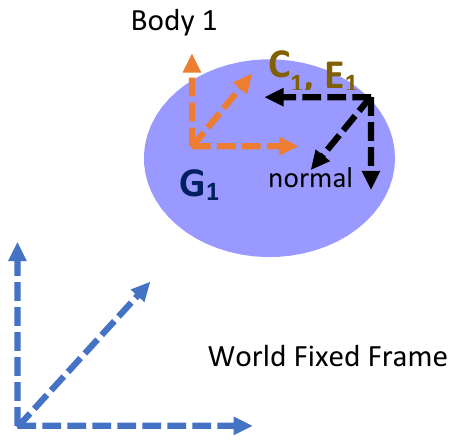}
     \caption{Frames attached to the object. The origin of frames $C_i$ and $E_i$ are coincident}
     \label{framesob1}
 \end{figure}
\section{Second Order Kinematics of Rolling}
\label{modelling}
\added{Consider two rigid bodies in contact}. For each rigid body, three associated frames are identified, as shown in Figure \ref{framesob1}. The frame {$G_i$} is attached, preferably to the geometric center of the body. The frame {$C_i$} is attached to the body and is located at the contact point on the surface of the rigid body in consideration. Let there be another frame ({$E_i$}), moving on the surface of the body, instantaneously coincident with the point of contact. The distance between the frames $C_i$ and $E_i$ is instantaneously equal to zero (by definition), but the relative velocity between the two frames may be non-zero. With the evolution of time, $E_i$ follows the instantaneous contact points (origin of instantaneous $C_i$ frame) synthesized on the surface of the body, describing a locus, which is the contact curve. The three frames ($C_i, G_i~ \text{and}~E_i$) can be assigned for each body in contact. The relative slip between the two bodies in contact is tracked by the instantaneous relative velocity between the $C_i$ frames attached to each body.

\begin{figure}[t]
    \centering
    \includegraphics[width=\textwidth, trim={2cm 18cm 2cm 2cm}]{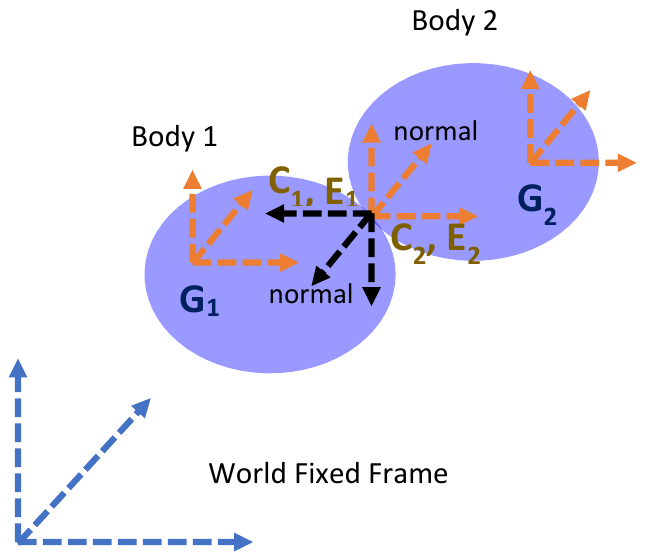}
    \caption{Frames attached to two objects. The origin of the frame $E_i$ generates the contact curves formed. The frame attached to $C_i$ not shown. The origin of frames $C_i$ and $E_i$ are coincident}
    \label{frames}
\end{figure}
\par In order to formulate the contact kinematics in terms of surface properties, the local surface coordinates are mapped to a cartesian system using localised chart functions.
The chart maps the surface coordinates of points on a surface to its cartesian coordinates. Let the position vector of the contact point ($\mathbf{r}$) be represented in terms of the local surface coordinates (u(s),v(s)) using the standard chart function \textbf{f}(u,v). The coordinate frame attached to $E_i$ is a natural coordinate frame defined by: $[\frac{\frac{\partial \textbf{f}}{\partial u}}{||\frac{\partial \textbf{f}}{\partial u}||},   \frac{\frac{\partial \textbf{f}}{\partial v}}{||\frac{\partial \textbf{f}}{\partial v}||},   \frac{\frac{\partial \textbf{f}}{\partial u}\times\frac{\partial \textbf{f}}{\partial v}}{||(\frac{\partial \textbf{f}}{\partial u}\times\frac{\partial \textbf{f}}{\partial v})||}]$. The velocity of the frame $E_i$ with respect to frame $G_i$ and written in the $G_i$ frame is given by equation \ref{velocitykin}.
\begin{equation}
    \label{velocitykin}
    \mathbf{v_{ei}} = \frac{d\mathbf{r}}{dt} = (\frac{\partial \mathbf{f}}{\partial u}{\dot{u}}+\frac{\partial \mathbf{f}}{\partial v}{\dot{v}})
\end{equation}
The rotation matrix relating the rotation of the $E_i$ frame is given by equation \ref{rotationmat}.
\begin{equation}
    \mathbf{R}^{G_i}_{E_i} = \left[\frac{\frac{\partial \mathbf{f}}{\partial u}}{||\frac{\partial \mathbf{f}}{\partial u}||},\frac{\frac{\partial \mathbf{f}}{\partial v}}{||\frac{\partial \mathbf{f}}{\partial v}||} ,\frac{\frac{\partial \mathbf{f}}{\partial u}\times\frac{\partial \mathbf{f}}{\partial v}}{||\frac{\partial \mathbf{f}}{\partial u}\times\frac{\partial \mathbf{f}}{\partial v}||}\right] = \left[{\frac{\partial \hat{\mathbf{f}}}{\partial u}},{\frac{\partial \hat{\mathbf{f}}}{\partial v}} ,{\hat{\mathbf{n}}}\right]
    \label{rotationmat}
\end{equation}
where $\hat{\mathbf{n}} = \frac{\frac{\partial \mathbf{f}}{\partial u}\times\frac{\partial \mathbf{f}}{\partial v}}{||\frac{\partial \mathbf{f}}{\partial u}\times\frac{\partial \mathbf{f}}{\partial v}||}$, $\mathbf{n} = {\frac{\partial {\mathbf{f}}}{\partial u}\times\frac{\partial {\mathbf{f}}}{\partial v}}$.
So, the angular velocity of the frame $E_i$ represented in frame $G_i$ is given by:
\begin{equation}
\label{angular}
    {\Omega}_{ei} = {{\mathbf{R}^{G_i}_{E_i}}}^T{\dot{\mathbf{R}}^{G_i}}_{E_i}
\end{equation}

\begin{equation}
  \text{with,}~  {\dot{\mathbf{R}}^{G_i}}_{E_i} = \dot{\mathbf{R_i}}\mathbf{K_i} + \mathbf{R_i}\dot{\mathbf{K_i}}
\end{equation}

where $\mathbf{R_i} = \left[{\frac{\partial \textbf{f}}{\partial u}},{\frac{\partial \textbf{f}}{\partial v}} ,{\frac{\partial \textbf{f}}{\partial u}}\times\frac{\partial \textbf{f}}{\partial v}\right]$ and $\mathbf{K_i}^{-1} = \left[\begin{array}{ccc}||\frac{\partial \textbf{f}}{\partial u}||&\textbf{0}&\textbf{0}\\\textbf{0}&||\frac{\partial \textbf{f}}{\partial v}||&\textbf{0}\\\textbf{0}&\textbf{0}&||\textbf{n}||\end{array}\right]$
\begin{equation}
        \dot{\mathbf{R_i}} = 
    \left[\frac{\partial ^2\mathbf{\hat{f}}}{\partial u^2}\dot{u}+\frac{\partial ^2\mathbf{\hat{f}}}{\partial u \partial v}\dot{v}, \frac{\partial ^2\mathbf{\hat{f}}}{\partial v^2}\dot{v}+\frac{\partial ^2\mathbf{\hat{f}}}{\partial v\partial u}\dot{u},\frac{\partial \mathbf{\hat{n}}}{\partial v}\dot{v}+\frac{\partial \mathbf{\hat{n}}}{\partial u}\dot{u}\right]
\end{equation}
The acceleration of the frame $E_i$ with respect to the frame $G_i$ is given by equation \ref{acc}.
\begin{equation}
\label{acc}
    \dot{\mathbf{v}}_{ei} = \left(\frac{\partial ^2\mathbf{f}}{\partial u^2}\dot{u}^2+\frac{\partial ^2\mathbf{f}}{\partial u\partial v}\dot{u}\dot{v}+\frac{\partial ^2\mathbf{f}}{\partial v \partial u}\dot{u}\dot{v}+\frac{\partial ^2\mathbf{f}}{\partial v^2}\dot{v}^2+\frac{\partial \mathbf{f}}{\partial u}{\ddot{u}}+\frac{\partial \mathbf{f}}{\partial v}{\ddot{v}}\right)
\end{equation}
The angular acceleration of the frame {$E_i$} is given by equation \ref{angacc}.
\begin{equation}
    \label{angacc}
    \dot{\mathbf{\Omega}}_{ei} = \mathbf{R}^{T}(\ddot{\mathbf{R_i}}\mathbf{K_i}+\dot{\mathbf{R_i}}{\mathbf{\dot{K}_i}}+\mathbf{R_i}\ddot{\mathbf{K_i}}+\dot{\mathbf{R_i}}{\mathbf{\dot{K}_i}})+\dot{\mathbf{R}}^T\dot{\mathbf{R}}
\end{equation}
Let $\frac{\partial \mathbf{\hat{f}}}{\partial u} = {\upzeta_1}$ and $\frac{\partial \mathbf{\hat{f}}}{\partial v} = \upzeta_2$, then the terms in equation \ref{angacc} take the form:

\begin{equation}
\begin{split}    
    \label{angacccomp}
    \ddot{\mathbf{R}}_i = [\frac{\partial ^2\upzeta_1}{\partial u^2}\dot{u}^2+\frac{\partial ^2\upzeta_1}{\partial u\partial v}\dot{u}\dot{v}+\frac{\partial \upzeta_1}{\partial u}\ddot{u}+\frac{\partial ^2\upzeta_1}{\partial v\partial u}\dot{u}\dot{v}+\frac{\partial ^2\upzeta_1}{\partial v^2}\dot{v}^2+\frac{\partial \upzeta_1}{\partial u}\ddot{v},\\ \frac{\partial ^2\upzeta_2}{\partial v^2}\dot{v}^2+\frac{\partial ^2\upzeta_2}{\partial v\partial u}\dot{v}\dot{u}+\frac{\partial \upzeta_2}{\partial v}\ddot{v}+\frac{\partial ^2\upzeta_2}{\partial u\partial v}\dot{v}\dot{u}+\frac{\partial^2\upzeta_2}{\partial u^2}\dot{u}^2+\frac{\partial \upzeta_2}{\partial v}\ddot{u} ]
\end{split}
\end{equation} 
We have thus derived the measures needed to write the kinematics equations, in cartesian frames. The kinematic equations are written here in terms of the surface coordinates and their time derivatives.
\section{Relative Kinematics of the Contacting Frames}
\label{finger}
\begin{figure}
    \centering
    \includegraphics[width=\textwidth, trim = {3cm 19cm 3cm 3cm}, clip]{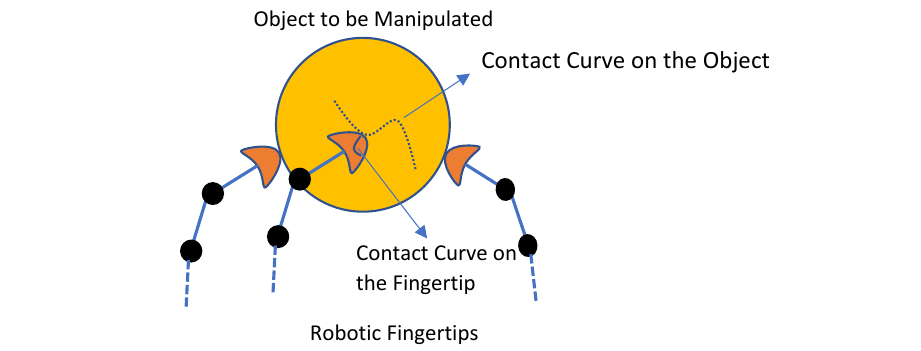}
    \caption{Example of in-hand manipulation. One of the fingers is shown to formulate a contact curve}
    \label{inhand}
\end{figure}

Consider a case of in-hand manipulation of the object using hard fingers (Figure \ref{inhand}). By hard fingers (\added{or rigid body}), we mean that the finger  has a representative point of contact. The process of \added{manipulation with the contacting body (or fingertip)} is desired to be a pure rolling motion. We assume that the contacting body (\added{or the fingertip}) does not spin about the local normal of the object at contact.
For robotics grasping, it is required that the fingers combine to apply forces to ensure that the contact between the object and the fingertip is always maintained. In this paper, we do not specifically look at means of ensuring contact and start by examining the interaction of the contact of the object with one of the fingers in isolation while the object is being manipulated.  \added{The interaction of the fingertip with the object can be generalized to any active rigid body (or manipulator end effector) interacting an object without break of contact. Throughout the derivation, the manipulating rigid body (fingertip) is referred to as body 1 whereas the the object is referred to as body 2.} To derive the expressions of relative acceleration and relative velocity using contact curves, the methods presented by Sankar et al. \cite{sankar1996velocity} are useful.  
\par The frames associated with body 1 and the object manipulated (body 2) is as shown in Figure \ref{frames}. In this formulation, the vector joining origin of frames  (say $A$ to say $B$) is represented as $\mathbf{r}_{AB} $. The velocity of a frame (let $B$) with respect to the frame (let $A$), as observed in frame (let $K$), is represented as ${}^K\mathbf{v}_{AB} $. Similarly, the acceleration of a frame (let $B$) with respect to the frame (let $A$), observed in frame (let $K$) is written as ${}^K\mathbf{a}_{AB} $ and, the angular velocity of frame (let $A$) as observed in frame (let $K$ ) is written as ${}^{K}\mathbf{\omega}_{A}$. \par
\begin{figure}[t]
    \centering
    \includegraphics[scale = 0.7, trim={2.5cm 12cm 0 2.5cm}, clip]{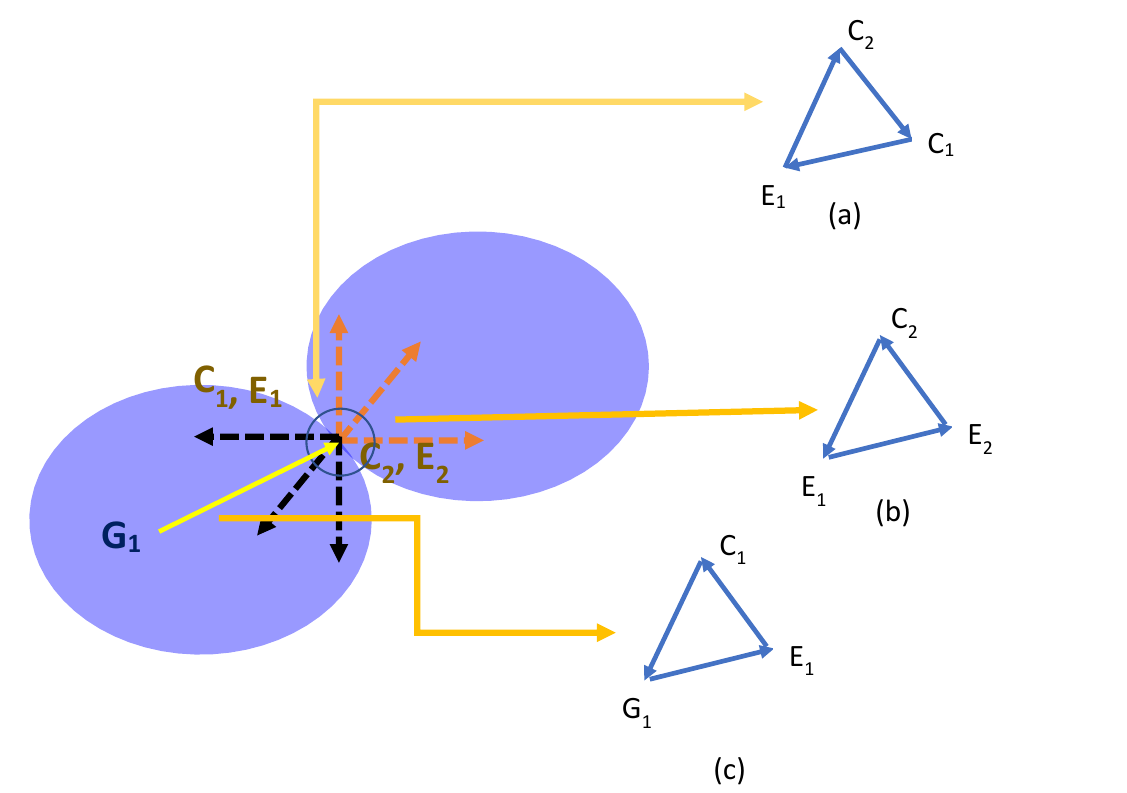}
    \caption{Virtual triangles of infinitesimal length connecting (a) $E_1 \rightarrow C_2 \rightarrow C_1 \rightarrow E_1$ (b)$E_1 \rightarrow E_2 \rightarrow C_2 \rightarrow E_1$. (c) shows the triangle connecting $G_1 \rightarrow E_1 \rightarrow C_1 \rightarrow G_1$ for body 1}
\label{triangle}
\end{figure}

 In order to relate the acceleration of the frames attached to respective bodies ($C_1$ and $C_2$), consider a triangle of infinitesimal sides at the contact point (connecting points $E_1~\text{to}~ C_2~\text{to}~C_1~\text{to}~E_1$) (Figure \ref{triangle} (a)). For these two contacting bodies, using triangle law of vector addition, we get:
\begin{equation}
\label{triangle_law}
    \mathbf{r}_{E_1C_2}+\mathbf{r}_{C_2C_1} +\mathbf{r}_{C_1E_1} = 0 
\end{equation}
Writing all the terms in equation \ref{triangle_law} in the frame $C_2$, and differentiating equation \ref{triangle_law} with respect to time in reference frame $C_2$ results in:
\begin{equation}
\label{velocitychange}
    {}^{C_2}\mathbf{v}_{E_1C_2}+{}^{C_2}\mathbf{v}_{C_2 C_1}+{}^{C_1}\mathbf{v}_{C_1 E_1} + {}^{C_2}\mathbf{\omega}_{C_1}\times\mathbf{r}_{C_1E_1} = 0
\end{equation}
We note here that $\mathbf{r_{C_1E_1}}$ is conveniently defined in frame $C_1$. Its derivative in frame $C_2$ consists of derivative of $\mathbf{r_{C_1E_1}}$ in frame $C_1$ and a term which is cross product of the relative angular velocity of the two frames with $\mathbf{r}_{C_1E_1}$. 
Differentiating once more in frame $C_2$, we relate the accelerations as:
\begin{equation}
    \label{acc1}
\begin{split}
   {}^{C_2}\mathbf{a}_{E_1C_2}+{}^{C_2}\mathbf{a}_{C_2C_1}+{}^{C_1}\mathbf{a}_{C_1E_1}+2{}^{C_2}\mathbf{\omega}_{C_1}\times{}^{C_1}\mathbf{v}_{C_1E_1}+\\{}^{C_2}\mathbf{\omega}_{C_1}\times({}^{C_2}\mathbf{\omega}_{C_1}\times\mathbf{r}_{C_1E_1})+{}^{C_2}\mathbf{\dot{\omega}}_{C_1}\times\mathbf{r}_{C_1E_1} = 0
\end{split}
\end{equation}
We also know that $\mathbf{r}_{C_1E_1}$ is equal to zero as the two frames are coincident by definition. However, a non-zero relative velocity between the frame $E_i$ and the frame $C_i$ is allowed. So, equation \ref{velocitychange} and equation \ref{acc1} reduce to:
\begin{equation}
\label{relvelacc}
\begin{split}
    {}^{C_2}\mathbf{v}_{C_2C_1} &= {}^{C_1}\mathbf{v}_{E_1C_1} -{}^{C_2}\mathbf{v}_{E_1C_2} \\
    {}^{C_2}\mathbf{a}_{C_2C_1} &= {}^{C_1}\mathbf{a}_{E_1C_1} -{}^{C_2}\mathbf{a}_{E_1C_2}-2{}^{C_2}\mathbf{\omega}_{C_1}\times{}^{C_2}\mathbf{v}_{C_1E_1}
\end{split}
\end{equation}
All the terms in equation \ref{relvelacc} are written in frame $C_2$. 
Similarly, for another case of triangle addition (Figure \ref{triangle} (b)), we get:
\begin{equation}
\label{another12}
    \mathbf{r}_{C_2E_1}+\mathbf{r}_{E_1E_2}+\mathbf{r}_{E_2C_2} = 0
\end{equation}
Again, differentiating the equation \ref{another12} twice with all the frames represented in $C_2$ yields:
\begin{equation}
    -{}^{C_2}\mathbf{v}_{E_1C_2}+{}^{E_2}\mathbf{v}_{E_1E_2}+ {}^{C_2}\mathbf{\omega}_{E_2}\times\mathbf{r}_{E_1E_2}+{}^{C_2}\mathbf{v}_{E_2 C_2} = 0
    \label{eqextra}
\end{equation}
As the contact frames $E_1$ and $E_2$ are by definition instantaneously coincident, so the relative velocity ${}^{E_2}\mathbf{v}_{E_1E_2} = 0~\text{and}~\mathbf{r}_{E_1E_2} = 0$ instantaneously. So, the equation \ref{eqextra} simplify into:
\begin{equation}
\begin{split}
   {}^{C_2}\mathbf{v}_{E_1C_2} = {}^{C_2}\mathbf{v}_{E_2C_2} 
\end{split}
\label{vextra}
\end{equation}
Similarly, differentiating equation \ref{eqextra} and substituting $\mathbf{v}_{E_1E_2} = 0, \mathbf{r}_{E_1E_2} = 0, {{}^{E_2}\mathbf{\dot{v}}_{E_1E_2}}=0$,$ \mathbf{\dot{r}}_{E_1E_2}=0$ we get equation \ref{accextra}. 
\begin{equation}
    {}^{C_2}\mathbf{a}_{E_1C_2} = {}^{C_2}\mathbf{a}_{E_2C_2}
    \label{accextra}
\end{equation}
Substituting LHS of equation \ref{vextra} and equation \ref{accextra} in equation \ref{relvelacc}, we get equation \ref{new}.
\begin{equation}
\begin{split}
    {}^{C_2}\mathbf{v}_{C_1C_2} &= -{}^{C_2}\mathbf{v}_{C_2C_1}= {}^{C_1}\mathbf{v}_{C_1E_1} -{}^{C_2}\mathbf{v}_{C_2E_2} = \mathbf{v}_{rel} \\
    {}^{C_2}\mathbf{a}_{C_1C_2} &= -{}^{C_2}\mathbf{a}_{C_2C_1} =   {}^{C_1}\mathbf{a}_{C_1E_1} -{}^{C_2}\mathbf{a}_{C_2E_2}+2{}^{C_2}\mathbf{\omega}_{C_1}\times{}^{C_2}\mathbf{v}_{C_1E_1} = \mathbf{a}_{rel}
\end{split}
\label{new}
\end{equation}
Through out the paper, all the terms are observed in the $C_2$ frame only (irrespective of the frame of representation). 
The angular velocity of the body 1 (frame $C_1$) as observed in frame $C_2$ is given by: $\mathbf{\omega}_{rel} = {}^{C_2}\mathbf{\omega}_{C_1}$. The relative angular velocity $\mathbf{\omega}_{rel}$ is then written in the $E_2$ frame as: $\mathbf{\omega}_{rel}=\left[ \omega_x~\omega_y~\omega_z\right]^T$, with $\omega_z$ directed along the local contact normal and $\omega_x$ and $\omega_y$ directed along orthogonal local tangents. With the stipulation that the objects in relative contact are not allowed to spin about the contact normal, the angular velocity term $\mathbf{\omega}_{rel}~  \text{reduces to}~ \left[ ~\omega_x~\omega_y~0~\right]^T$. From yet another case of triangle law of vector addition (Figure \ref{triangle} (c)), we have:
\begin{equation}
\label{another}
    \mathbf{r}_{G_iE_i} = \mathbf{r}_{G_iC_i}+\mathbf{r}_{C_iE_i}
\end{equation}
Differentiating equation \ref{another} in $G_i$ frame results to:
\begin{equation}
    {}^{G_i}\mathbf{v}_{G_iE_i} = {}^{G_i}\mathbf{v}_{G_iC_i}+{}^{C_i}\mathbf{v}_{C_iE_i}+{}^{G_i}\mathbf{\omega}_{C_i}\times\mathbf{r}_{C_iE_i}
\end{equation}
As, the frames $C_i$ and $G_i$ are on the same rigid body, so ${}^{G_i}\mathbf{v}_{G_iE_i} = {}^{C_i}\mathbf{v}_{C_iE_i} (\mathbf{r}_{C_iE_i} = 0, {}^{G_i}\mathbf{v}_{G_iC_i} = 0)$. By a similar argument, ${}^{G_i}\mathbf{a}_{G_iE_i} = {}^{C_i}\mathbf{a}_{C_iE_i}$. Hence, the acceleration and velocity of the frame $E_i$ with respect to $G_i$ as observed in $G_i$ are equal to the acceleration and velocity of the frame $E_i$ with respect to $C_i$, as observed in $C_i$. We can state this equation form as: 
\begin{equation}
\begin{split}
   {}^{C_i}\mathbf{v}_{C_iE_i} = {}^{G_i}\mathbf{v}_{G_iE_i} =  \mathbf{v}_{ei} \\
   {}^{C_i}\mathbf{a}_{C_iE_i} =
    {}^{G_i}\mathbf{a}_{G_iE_i} =  \mathbf{a}_{ei}
    \end{split}
\end{equation}
In order to formulate the relative velocity or acceleration, it is important to write all the terms in the same frame (irrespective to the frame of observation). All the terms are written in the $E_2$ frame. Appropriate rotation matrices are used to change the frame of representation of vectors.
Following this method, the relative acceleration between the frames $C_1$ and $C_2$ (equation \ref{new}), written in a frame attached to $E_2$ are obtained by premultiplying the rotation matrices.

\begin{equation}
    \label{relacc}
    \mathbf{a}_{rel} = \mathbf{R}_1\dot{v}_{e1}-\mathbf{R}_2\dot{v}_{e2}+2\mathbf{\omega}_{rel}\times\mathbf{R}_1\mathbf{v}_{e1}
\end{equation}
where $\mathbf{R}_1$ and $\mathbf{R}_2$ represents the rotation matrices required to transform the acceleration terms from $G_i$ frames to one of the contact frames (here $E_2$). For this case, the rotation matrices $\mathbf{R}_1$ and $\mathbf{R}_2$ are:
\begin{equation}
    \label{rotation}
    \mathbf{R}_1 = \mathbf{R}_{\psi}^{T}\mathbf{R}^{G_1^{T}}_{E_1}, \mathbf{R}_2 = \mathbf{R}^{G_2^T}_{E_2} 
\end{equation}
where $\mathbf{R}_{\psi} = \mathbf{R}_{\psi}^T = \left[\begin{array}{ccc}cos\psi&-sin\psi&0\\-sin\psi&-cos\psi&0\\0&0&-1\end{array}\right]$, and $\psi$ is the angle made by the $\frac{\mathbf{\partial f_i}}{\partial u_i}$ vectors corresponding to each contact curve on the two surfaces. The details of the syntesis of the rotation matrix ($\mathbf{R}_{\psi}$ for two contacting bodies is explained in detail in the literature \cite{montana1988kinematics, han1997dextrous}). The matrix $\mathbf{R}_{\psi}$ is used to relate vectors written in $E_1$ frame to $E_2$ frame. The angle $\psi$ refers to the angle between the x-axis (u-curves) for the two contacting bodies. 
\par Substituting $\frac{\partial \hat{\mathbf{f_i}}^T}{\partial u_i}\frac{\partial {\hat{\mathbf{f_i}}}}{\partial u_i} = 1$,$\frac{\partial \mathbf{\hat{f_i}}^T}{\partial v_i}\frac{\partial \mathbf{\hat{f_i}}}{\partial v_i} = 1$ and $\frac{\partial \hat{\mathbf{f_i}}^T}{\partial u_i}\frac{\partial \hat{\mathbf{f_i}}}{\partial v_i} = 0$ and by using the definitions of second order Christoffel symbols (equation \ref{christoffel}), we simplify the expression of relative acceleration terms. The elements $\frac{\partial \hat{\mathbf{f_i}}^T}{\partial u_i}\frac{\partial \hat{\mathbf{f_i}}}{\partial v_i}$ are the non diagonal terms of the metric tensor \cite{murray2017mathematical}  of the surface. It is known that the natural coordinate system (u,v) is orthogonal. So by definition, the non diagonal elements of the metric tensor are equal to zero, that is, $\frac{\partial \mathbf{f_i}}{\partial u_i}^T\frac{\partial \mathbf{f_i}}{\partial v_i} = 0$. So, the terms corresponding to the non diagonal elements of the metric tensor drop out of the formulation. Using the more compact resulting Christoffel symbols:
\begin{equation}
\label{christoffel}
\begin{split}
    {\Gamma_{11}^{1}}_i = \frac{\partial {\mathbf{f}_i}^T}{\partial u_i}\frac{\partial ^2{\mathbf{f}_i}}{\partial u_i^2}\frac{1}{||\frac{\partial \mathbf{f_i}}{\partial u_i}||^2},    {\Gamma_{12}^{1}}_i = \frac{\partial {\mathbf{f}_i}^T}{\partial u_i}\frac{\partial ^2{\mathbf{f}_i}}{\partial u_i\partial v_i}\frac{1}{||\frac{\partial \mathbf{f_i}}{\partial u_i}||^2},  {\Gamma_{22}^{1}}_i = \frac{\partial {\mathbf{f}_i}^T}{\partial u_i}\frac{\partial ^2{\mathbf{f}_i}}{\partial v_i^2}\frac{1}{||\frac{\partial \mathbf{f_i}}{\partial u_i}||^2} \\
    {\Gamma_{11}^{2}}_i = \frac{\partial {\mathbf{f_i}}^T}{\partial v_i}\frac{\partial ^2{\mathbf{f}_i}}{\partial u_i^2}\frac{1}{||\frac{\partial \mathbf{f_i}}{\partial v_i}||^2},    {\Gamma_{12}^{2}}_i = \frac{\partial {\mathbf{f}_i}^T}{\partial v_i}\frac{\partial ^2{\mathbf{f}_i}}{\partial u_i\partial v_i}\frac{1}{||\frac{\partial \mathbf{f_i}}{\partial v_i}||^2},  {\Gamma_{22}^{2}}_i = \frac{\partial {\mathbf{f}_i}^T}{\partial v_i}\frac{\partial ^2{\mathbf{f}_i}}{\partial v_i^2}\frac{1}{||\frac{\partial \mathbf{f_i}}{\partial v_i}||^2}
\end{split}
\end{equation}
The equation for relative acceleration in equation \ref{relacc} is rewritten as equation \ref{finetune}.
\begin{equation}
    \label{finetune}
\begin{split}
\mathbf{a}_{rel} = \mathbf{R}_{\psi}\mathbf{M_1}\underbrace{\left[\begin{array}{ccc}\ddot{u}_1+{\Gamma_{11}^{1}}_1\dot{u}_1^2+2{\Gamma_{12}^{1}}_1\dot{u}_1\dot{v}_1+{\Gamma_{22}^{1}}_1\dot{v}_1^2 \\ \ddot{v}_1+{\Gamma_{11}^{2}}_1\dot{u}_1^2+2{\Gamma_{12}^{2}}_1\dot{u}_1\dot{v}_1+{\Gamma_{22}^{2}}_1\dot{v}_1^2 \\ * \end{array}\right]}_{\mathbf{a_1}_{E_1}} - \\\mathbf{M_2}\underbrace{\left[\begin{array}{ccc}\ddot{u}_2+{\Gamma_{11}^{1}}_2\dot{u}_2^2+2{\Gamma_{12}^{1}}_2\dot{u}_2\dot{v}_2+{\Gamma_{22}^{1}}_2\dot{v}_2^2 \\ \ddot{v}_2+{\Gamma_{11}^{2}}_2\dot{u}_2^2+2{\Gamma_{12}^{2}}_2\dot{u}_2\dot{v}_2+{\Gamma_{22}^{2}}_2\dot{v}_2^2 \\ * \end{array}\right]}_{\mathbf{a_2}_{E_2}}
\end{split}
\end{equation}
where $\mathbf{M_i}$ is given by $\mathbf{M_i} = \left[\begin{array}{ccc}||\frac{\partial \mathbf{f_i}}{\partial u_i}||&0&0 \\ 0&||\frac{\partial \mathbf{f_i}}{\partial v_i}||&0\\0&0&1\end{array}\right]$.  The first two rows represent the tangential direction and the last element represents the acceleration along the normal direction. In order to ensure brevity, the normal acceleration component is not derived explicitly and labelled as `*'.  
Similarly, the relative velocity can be written as equation \ref{relvel}.
\begin{equation}
    \label{relvel}
    \mathbf{v}_{rel} = \mathbf{R_1}\mathbf{v}_{e1}-\mathbf{R_2}\mathbf{v}_{e2}
\end{equation}
Components of the relative velocity can be obtained by substituting the expression of rotation matrices and expression of $\mathbf{v}_{ei}$ as in equation \ref{velocitykin} in equation \ref{relvel}.
\begin{equation}
\label{netrelvel}
\mathbf{v}_{rel} =  \left[\begin{array}{c}||\frac{\partial \mathbf{f_1}}{\partial u_1}||\dot{u}_1cos\psi-||\frac{\partial \mathbf{f_1}}{\partial v_1}||\dot{v}_1sin\psi-||\frac{\partial \mathbf{f_2}}{\partial u_2}||\dot{u}_2 \\ -||\frac{\partial \mathbf{f_1}}{\partial u_1}||\dot{u}_1sin\psi-||\frac{\partial \mathbf{f_1}}{\partial v_1}||\dot{v}_1cos\psi-||\frac{\partial \mathbf{f_2}}{\partial v_2}||\dot{v}_2 \\ 0
    \end{array}\right]
\end{equation}
The first two rows represent the velocities in the tangent plane to the surface at the contact point to the object. In order to restrict slipping, we need to ensure that the first two terms in the two rows in equation \ref{finetune} are zero. \par
In order to plan the trajectory, it is useful to have a single variable parameterisation of the path parameters ($u_i(s), v_i(s)$) with respect to time. The rate of growth of the path parameter `s' is related to time as $\sigma = \frac{ds}{dt}$. For a particular manipulation, the term $\sigma$ can be planned to be a time-varying polynomial. The rate of growth of path parameters $(u_i(s),v_i(s))$ with respect to time is represented by equation \ref{sigma}.
\begin{equation}
    \label{sigma}
    \dot{u_i} = \sigma{\frac{du_i}{ds}}, \dot{v_i} = \sigma\frac{dv_i}{ds}
\end{equation}
On differentiating equation \ref{sigma}, we get:
\begin{equation}
    \label{timesigma}
    \ddot{u_i} = \dot{\sigma}\frac{du_i}{ds}+\sigma^2\frac{d^2u_i}{ds^2},\ddot{v_i} = \dot{\sigma}\frac{dv_i}{ds}+\sigma^2\frac{d^2v_i}{ds^2} 
\end{equation}
The expressions in equation \ref{timesigma} are reformulated to obtain second order path derivatives:
\begin{equation}
\begin{split}
    \label{timesigmareform}
    \frac{d^2u_i}{ds^2} = (\ddot{u_i}-\frac{\dot{\sigma}}{\sigma}\dot{u_i})\frac{1}{\sigma^2} \\
       \frac{d^2v_i}{ds^2} = (\ddot{v_i}-\frac{\dot{\sigma}}{\sigma}\dot{v_i})\frac{1}{\sigma^2}
    \end{split}
\end{equation}
The equation \ref{sigma} and equation \ref{timesigmareform} relate the first and second time derivative of the growth of surface coordinates to the first and second derivatives with respect to the path parameter `s'. This reformulation allows computation of geodesics in terms of time derivative of the surface coordinates. 
The geodesic curve is commonly expressed in differential form \cite{haw1983geodesic} in terms of the path parameter `s':
\begin{equation}
\begin{split}
\label{geodesic}
\frac{d^2u_i}{ds^2} + {{\Gamma_{11}^{1}}_i}(\frac{du_i}{ds})^2+2{{\Gamma_{12}^{1}}_i}\frac{du_i}{ds}\frac{dv_i}{ds}+{{\Gamma_{22}^{1}}_i}(\frac{dv_i}{ds})^2 = 0   \\
\frac{d^2v_i}{ds^2} + {{\Gamma_{11}^{2}}_i}(\frac{du_i}{ds})^2+2{{\Gamma_{12}^{2}}_i}\frac{du_i}{ds}\frac{dv_i}{ds}+{{\Gamma_{22}^{2}}_i}(\frac{dv_i}{ds})^2 = 0
\end{split}
\end{equation}
The equation \ref{geodesic} is written in terms of the derivatives of the surface coordinates with respect to the path parameters. To obtain the geodesic equations in terms of the derivatives of the contact coordinates with respect to time, the terms relating $\frac{du_i}{ds},\frac{dv_i}{ds}, \frac{d^2u_i}{ds^2}, \frac{d^2v_i}{ds^2}$ (in equation \ref{sigma} and \ref{timesigmareform}) are substituted in equation \ref{geodesic}. The time dependent version of equation \ref{geodesic} results in equation \ref{timegeodesic}.

\begin{equation}
\begin{split}
\label{timegeodesic}
\frac{1}{\sigma^2}\ddot{u}_i-\frac{\dot{\sigma}}{\sigma^3}\dot{u}_i +{{\Gamma_{11}^{1}}_i}\frac{1}{\sigma^2}\dot{u}_i^2+2{{\Gamma_{12}^{1}}_i}\frac{1}{\sigma^2}\dot{u}_i\dot{v}_i+{{\Gamma_{22}^{1}}_i}\frac{1}{\sigma^2}\dot{v}_i^2 = 0 \\
\frac{1}{\sigma^2}\ddot{v}_i-\frac{\dot{\sigma}}{\sigma^3}\dot{v}_i +{{\Gamma_{11}^{2}}_i}\frac{1}{\sigma^2}\dot{v}_i^2+2{{\Gamma_{12}^{2}}_i}\frac{1}{\sigma^2}\dot{u}_i\dot{v}_i+{{\Gamma_{22}^{2}}_i}\frac{1}{\sigma^2}\dot{v}_i^2 = 0
\end{split}
\end{equation}
We shall call the reformulated governing equation \ref{timegeodesic} as a \textit{time parameterized geodesic trajectory}.
Substitute expressions of $\ddot{u}$ and $\ddot{v}$ (from equation \ref{timegeodesic}) in equation \ref{relacc}, to get an expression of $a_{rel}$ as in equation \ref{final}.

\begin{equation}
\label{final}
    \mathbf{a}_{rel} = \left[\begin{array}{ccc}||\frac{\partial \mathbf{f_1}}{\partial u_1}||n_1cos\psi-||\frac{\partial \mathbf{f_1}}{\partial v_1}||n_2sin\psi-||\frac{\partial \mathbf{f_2}}{\partial u_2}||n_3\\ -||\frac{\partial \mathbf{f_1}}{\partial u_1}||n_1sin\psi - ||\frac{\partial \mathbf{f_1}}{\partial v_1}||n_2cos\psi - ||\frac{\partial \mathbf{f_2}}{\partial v_2}||n_4 \\ *\end{array}\right]
\end{equation}
where $n_1$, $n_2$, $n_3$ and $n_4$ are defined by:
\begin{equation}
\label{n1234}
\begin{split}
n_1 = \frac{\dot{\sigma}}{\sigma}\dot{u}_1 -{{\Gamma_{11}^{1}}_1}\dot{u}_1^2-2{{\Gamma_{12}^{1}}_1}\dot{u}_1\dot{v}_1-{{\Gamma_{22}^{1}}_1}\dot{v}_1^2+{\Gamma_{11}^{1}}_1\dot{u}_1^2+2{\Gamma_{12}^{1}}_1\dot{u}_1\dot{v}_1+{\Gamma_{22}^{1}}_1\dot{v}_1^2 = \frac{\dot{\sigma}}{\sigma}\dot{u}_1 \\
n_2 = \frac{\dot{\sigma}}{\sigma}\dot{v}_1 -{{\Gamma_{11}^{2}}_1}\dot{v}_1^2-2{{\Gamma_{12}^{2}}_1}\dot{u}_1\dot{v}_1-{{\Gamma_{22}^{2}}_1}\dot{v}_1^2+{\Gamma_{11}^{2}}_1\dot{u}_1^2+2{\Gamma_{12}^{2}}_1\dot{u}_1\dot{v}_1+{\Gamma_{22}^{2}}_1\dot{v}_1^2 = \frac{\dot{\sigma}}{\sigma}\dot{v}_1\\
n_3 = \frac{\dot{\sigma}}{\sigma}\dot{u}_2 -{{\Gamma_{11}^{1}}_2}\dot{u}_2^2-2{{\Gamma_{12}^{1}}_2}\dot{u}_2\dot{v}_2-{{\Gamma_{22}^{1}}_2}\dot{v}_2^2+{\Gamma_{11}^{1}}_2\dot{u}_2^2+2{\Gamma_{12}^{1}}_2\dot{u}_2\dot{v}_2+{\Gamma_{22}^{1}}_2\dot{v}_2^2 = \frac{\dot{\sigma}}{\sigma}\dot{u}_2 \\
n_4 = \frac{\dot{\sigma}}{\sigma}\dot{v}_2 -{{\Gamma_{11}^{2}}_2}\dot{v}_2^2-2{{\Gamma_{12}^{2}}_2}\dot{u}_2\dot{v}_2-{{\Gamma_{22}^{2}}_2}\dot{v}_2^2+{\Gamma_{11}^{2}}_2\dot{u}_2^2+2{\Gamma_{12}^{2}}_2\dot{u}_2\dot{v}_2+{\Gamma_{22}^{2}}_2\dot{v}_2^2 = \frac{\dot{\sigma}}{\sigma}\dot{v}_2
\end{split}
\end{equation}
and, equivalently equation \ref{newfinal}.
\begin{equation}
    \label{newfinal}
    \mathbf{a}_{rel} = \left[\begin{array}{ccc}||\frac{\partial \mathbf{f_1}}{\partial u_1}||\frac{\dot\sigma}{\sigma}\dot{u}_1cos\psi-||\frac{\partial \mathbf{f_1}}{\partial v_1}||\frac{\dot\sigma}{\sigma}\dot{v}_1sin\psi-||\frac{\partial \mathbf{f_2}}{\partial u_2}||\frac{\dot{\sigma}}{\sigma}\dot{u}_2 \\ -||\frac{\partial \mathbf{f_1}}{\partial u_1}||\frac{\dot\sigma}{\sigma}\dot{u}_1sin\psi-||\frac{\partial \mathbf{f_1}}{\partial v_1}||\frac{\dot\sigma}{\sigma}\dot{v}_1cos\psi-||\frac{\partial \mathbf{f_2}}{\partial v_2}||\frac{\dot{\sigma}}{\sigma}\dot{v}_2 \\ *
     \end{array}\right]
\end{equation}
We see that the expression $\mathbf{a}_{rel}$ is a scalar multiple of the expression of $\mathbf{v}_{rel}$ in equation \ref{netrelvel} and can be conveniently written as:
\begin{equation}
\label{velacceleration}
    \mathbf{a}_{rel} =  \frac{\dot{\sigma}}{\sigma}\left[\begin{array}{ccc}{v_{rel}}_x \\ {v_{rel}}_y \\ *\end{array}\right]
\end{equation}
The equation \ref{velacceleration} suggests that the time parameterized geodesic based trajectory of the contact point on the \added{two contacting bodies} leads to a relative acceleration which is a scalar multiple of the relative tangential velocity between the object and the finger. For the case of the relative velocity being zero at the start of the manipulation, the relative acceleration between the contacts will be zero for all subsequent time instances. Also, in case the rate of change of the position of the contact point with path parameter on the object being identical to or a constant scalar multiple of the time rate of change of the contact point ($\sigma = \text{constant}, \dot{\sigma}=0$), then a reformulation of equation \ref{n1234}, results in $n_i = 0$ for i = 1:4. This suggests that the relative acceleration \added{of the origin of the two contacting frames is equal to zero}. If $\sigma = \text{constant}$, then the equation \ref{velacceleration} simplifies to: 
\begin{equation}
\label{important}
\mathbf{a}_{rel} = \left[\begin{array}{ccc}0\\0\\ *\\ \end{array}\right] 
\end{equation}
 
\par To interpret equation \ref{important}, consider a second order rolling condition, which is satisfied between two bodies in contact and relative motion if both the relative velocity and the relative acceleration of the instantaneous contact point is zero. 
If no slip is ensured for finite time $\delta t$, that is, if $v_{rel}$ = 0 over a time interval $\delta t$, then the relative tangential acceleration is zero at all times, ensuring second order rolling between the finger and the object.
As a particular case, if $\sigma$ is planned to be a zero-order polynomial (constant value), that is, $\dot{\sigma}=0$, then the contact should necessarily be initiated contact with an initial relative velocity. In case the initial relative velocity at contact is zero, the contacting bodies will not move each other. So, this is not an attractive option. Hence, a proper design requires $\sigma$ to be time-varying entity, perhaps a higher-order polynomial. \par
In order to prevent slip during manipulation, the locus of the contact point should map geodesic curves on the fingertip ({body 1}), and conversely, the contact curve on the object  should be geodesic as well. We note that the proof has been derived for exact geodesic trajectories. Though conceptually simple, the method is limited as it requires that the relative velocity between the contacting bodies to be maintained at zero for a small but finite time $\delta t$. This method also requires that the contact trajectory strictly follows the  based contact curve at all time instances. We hence have no means of proper disturbance rejection if the contact trajectory deviates from the geodesic. It is understood that a practical system will be realisable if the contact curves are natural attractors, thus enabling proper disturbance rejection. In order to ensure proper disturbance rejection, the underlying geodesic trajectories are modified (section \ref{dist}).   

\section{Disturbance Rejection}
\label{dist}
In order to ensure proper disturbance rejection for the trajectories, that is, in the case the contact system acquires relative velocity (sliding), then the trajectory should be controlled such that the relative velocity tends to zero asymptotically. This can be done by modifying the relationship (equation \ref{velacceleration}) between $\mathbf{a}_{rel}$ and $\mathbf{v}_{rel}$ in equation \ref{velacceleration} to the form in equation \ref{contracaccel}.

\begin{equation}
    \label{contracaccel}
    a_{rel} =  \{\frac{\dot{\sigma}}{\sigma}-\eta\sigma^2\}\left[\begin{array}{ccc}{v_{rel}}_x \\ {v_{rel}}_y \\ *\end{array}\right]
\end{equation}
where $\eta$ is a positive scalar. A selection of $\eta$ ensures that the system is in a contraction region \cite{lohmiller1998contraction} with $\frac{\dot{\sigma}}{\sigma}-\eta\sigma^2$ being negative over the entire trajectory.
 Correspondingly, the  governing equations for contact curves are modified from equation \ref{timegeodesic} to equation \ref{contrac}. The trajectory on body 1 (fingertip) remains the same as earlier. Only the equations governing the geodesic curve mapped on the object (body 2) change as a result of this modification. 
\begin{equation}
\begin{split}
\frac{1}{\sigma^2}\ddot{u}_1-\frac{\dot{\sigma}}{\sigma^3}\dot{u}_1 +{{\Gamma_{11}^{1}}_1}\frac{1}{\sigma^2}\dot{u}_1^2+2{{\Gamma_{12}^{1}}_1}\frac{1}{\sigma^2}\dot{u}_1\dot{v}_1+{{\Gamma_{22}^{1}}_1}\frac{1}{\sigma^2}\dot{v}_1^2 = 0 \\
\frac{1}{\sigma^2}\ddot{v}_1-\frac{\dot{\sigma}}{\sigma^3}\dot{v}_1 +{{\Gamma_{11}^{2}}_1}\frac{1}{\sigma^2}\dot{v}_1^2+2{{\Gamma_{12}^{2}}_1}\frac{1}{\sigma^2}\dot{u}_1\dot{v}_1+{{\Gamma_{22}^{2}}_1}\frac{1}{\sigma^2}\dot{v}_1^2 = 0 \\
\frac{1}{\sigma^2}\ddot{u}_2-\frac{\dot{\sigma}}{\sigma^3}\dot{u}_2 +{{\Gamma_{11}^{1}}_2}\frac{1}{\sigma^2}\dot{u}_2^2+2{{\Gamma_{12}^{1}}_2}\frac{1}{\sigma^2}\dot{u}_2\dot{v}_2+{{\Gamma_{22}^{1}}_2}\frac{1}{\sigma^2}\dot{v}_2^2 = \\ \eta\left(||\frac{\partial \mathbf{f_2}}{\partial u_2}||^{-1}\{||\frac{\partial \mathbf{f_1}}{\partial u_1}||\dot{u}_1cos(\psi)-||\frac{\partial \mathbf{f_1}}{\partial v_1}||\dot{v}_1sin(\psi)\}-\dot{u}_2\right) \\
\frac{1}{\sigma^2}\ddot{v}_2-\frac{\dot{\sigma}}{\sigma^3}\dot{v}_2 +{{\Gamma_{11}^{2}}_2}\frac{1}{\sigma^2}\dot{v}_2^2+2{{\Gamma_{12}^{2}}_2}\frac{1}{\sigma^2}\dot{u}_2\dot{v}_2+{{\Gamma_{22}^{2}}_2}\frac{1}{\sigma^2}\dot{v}_2^2 \\= \eta\left(||\frac{\partial \mathbf{f_2}}{\partial v_2}||^{-1}\{-||\frac{\partial \mathbf{f_1}}{\partial u_1}||\dot{u}_1sin(\psi)-||\frac{\partial \mathbf{f_1}}{\partial v_1}||\dot{v}_1cos(\psi)\}-\dot{v}_2)\right)
\end{split}
\label{contrac}
\end{equation}
 Contacts satisfying equation \ref{contrac}, ensure disturbance rejection if the fingertips (body 1) instantaneously slip on the object (body 2) or a lateral disturbance is forced by the environment. The modified geodesic equations can be used to define the contact curve at the next instance $\left\{(u_1,v_1), (\dot{u}_1, \dot{v}_1), (u_2, v_2), (\dot{u}_2, \dot{v}_2)\right\}$ when the contact is known during relative slip. The right hand side terms of the third and the fourth equation grow to non zero values when the contacting frames show relative slip. The modified growth of contact curves on the surface of the object (or body 2) lead to rejection of slip. The modified geodesic trajectories converge to being exact geodesic trajectories once the relative velocity goes to zero. 
\subsection{Numerical example with a sphere}
\label{sphere}
Consider the case of three spherical fingers manipulating a spherical object. For each object - fingertip interface, the body 2 contact variables $(u_2, v_2)$ are represented as the object contact curve at the $i^{th}$ contact $(u_{oi}, v_{oi})$ and the body 1 contact variables are represented as the fingertip contact variables $(u_{fi}, v_{fi})$. Let the local chart variables on the $i^{th}$ contact on the object be $(u_{oi},v_{oi})$ and on the $i^{th}$ finger be $(u_{fi},v_{fi})$. The chart for the $k^{th}$ sphere is given by:
\begin{equation}
\mathbf{f_k}(u_k, v_k) = \left[\begin{array}{c}r_ksin(u_k)cos(v_k)\\~r_ksin(u_k)sin(v_k)\\ ~r_kcos(u_k)\end{array}\right]
\end{equation}
The direction of the axes of the local frames $E_k$ (the frame moving on the object surface for each of the spherical body) is given by equation \ref{spherech}.

\begin{equation}
\begin{split}
    \label{spherech}
    \frac{\partial \mathbf{f_k}}{\partial u_k} &= e_x = \{r_kcos(u_k)cos(v_k),r_kcos(u_k)sin(v_k),-r_ksin(u_k)\} \\
    \frac{\partial \mathbf{f_k}}{\partial v_k} &= e_y = \{-r_ksin(u_k)sin(v_k),r_ksin(u_k)cos(v_k),0\} \\
    \mathbf{\hat{n}_k} &= e_z = \{sin(u_k)cos(v_k),sin(u_k)sin(v_k),cos(u_k)\} \\
\end{split}
\end{equation}
where $\mathbf{f_k}$ is the chart, $r_k$ is the radius, $(u_k,v_k)$ are the contact coordinates for the $k^{th}$ sphere. For this example, we use the subscript ($oi$) for the object and the subscript ($f_i$) for the fingers, with chart functions being the same. 
\begin{figure}[t]
    \centering
    \includegraphics[scale = 0.8, trim={2.5cm 12cm 0 2.5cm}, clip]{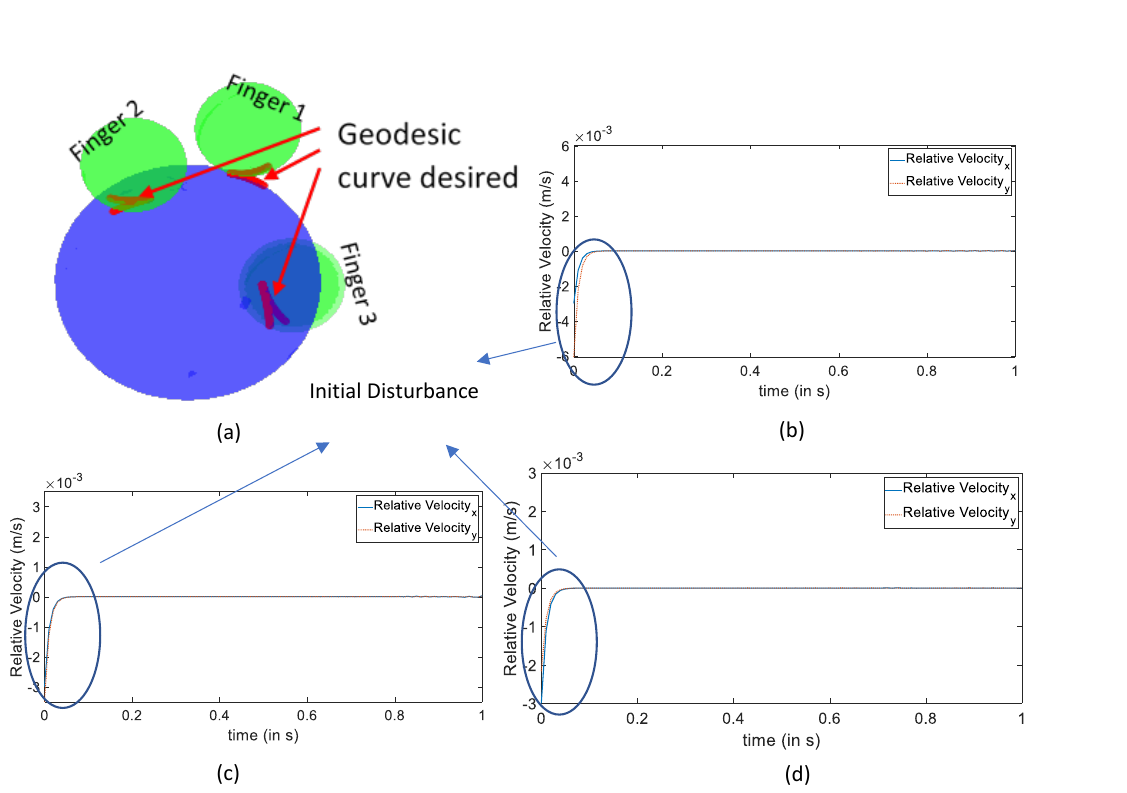}
    \caption{(a) The finger curves and the curve on the object generated by the acceleration based geodesic equation. Relative velocity in the local x-y frame for (a) finger 1 - object, (c) finger 2 - object and (c) finger 3 - object}
\label{basic}
\end{figure}
\begin{figure}[t]
    \centering
    \includegraphics[scale = 0.6, trim={3cm 1cm 0 2cm}, clip]{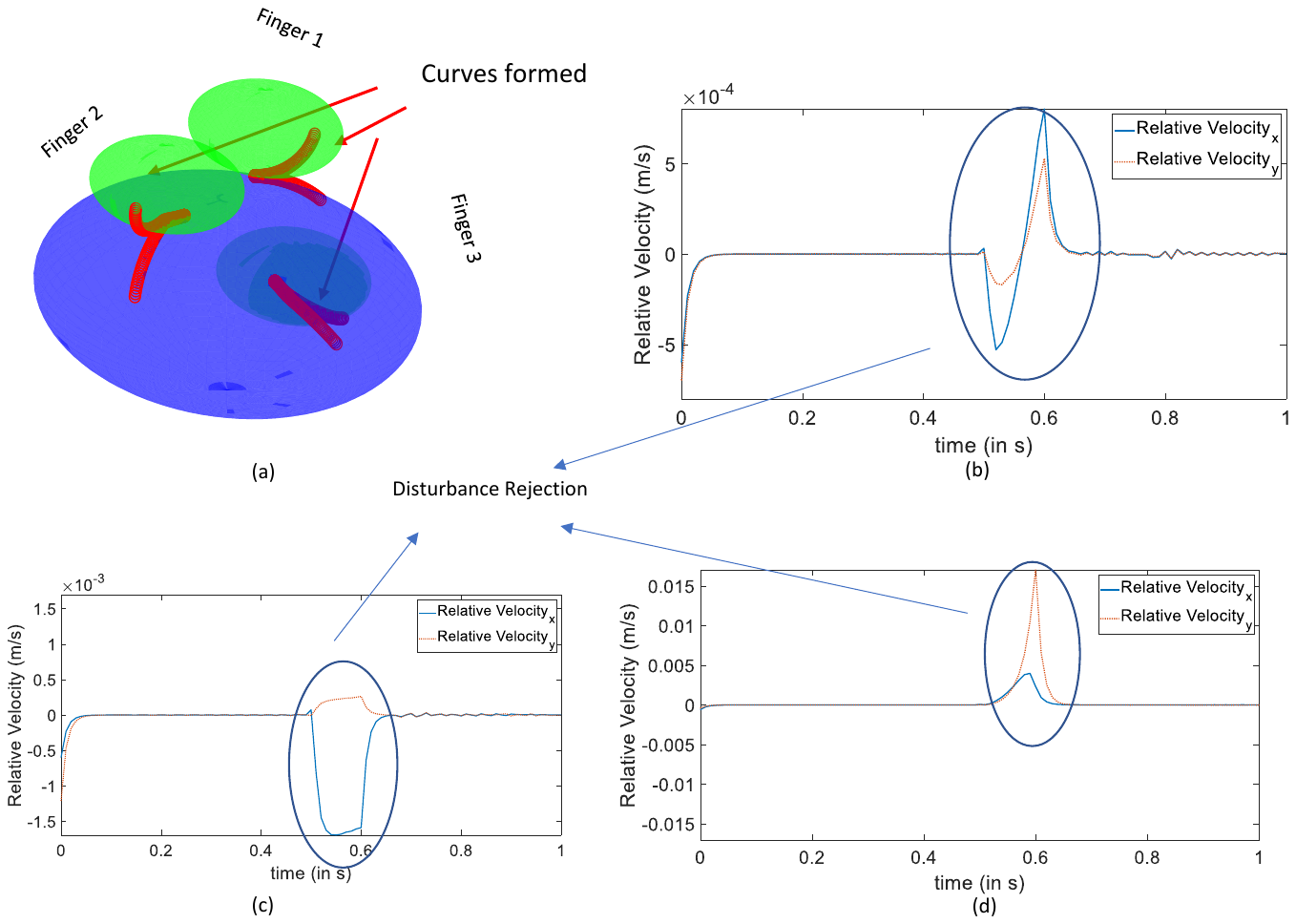}
    \caption{(a) The curves formed on the finger and the object. (b) Relative velocity between object and finger - 1, (c) for object and finger - 2 (d) for object and finger - 3}
    \label{contract}
\end{figure}

\begin{figure}[t]
    \centering
    \includegraphics[scale = 0.8, trim = {2.2cm 9.5cm 0 2.5cm}, clip]{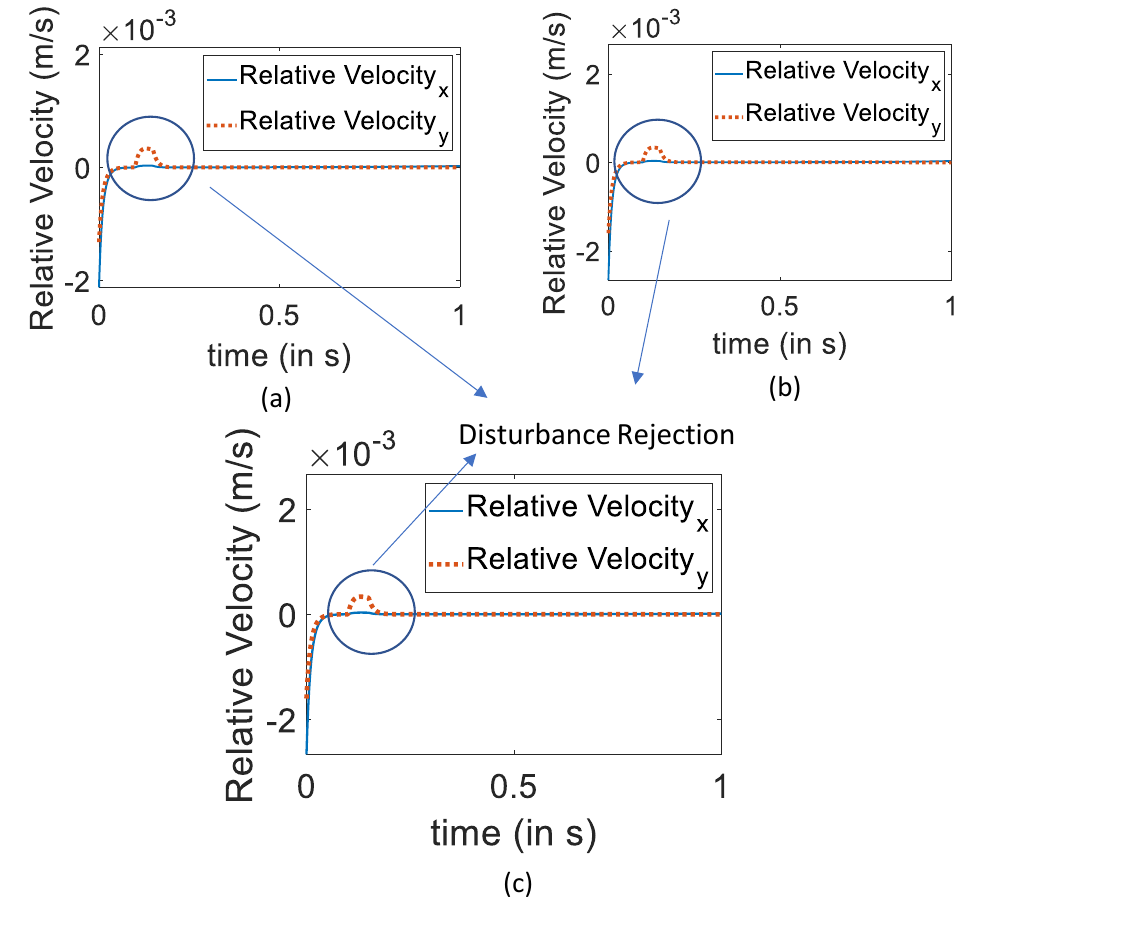}
    \caption{Relative velocity at the finger - object interface for (a) First finger (b) Second Finger (c) Third Finger}
    \label{ellipsoid_result}
\end{figure}
\begin{figure}[t]
    \centering
    \includegraphics[scale = 0.8, trim = {2.2cm 17cm 0 2.5cm}, clip]{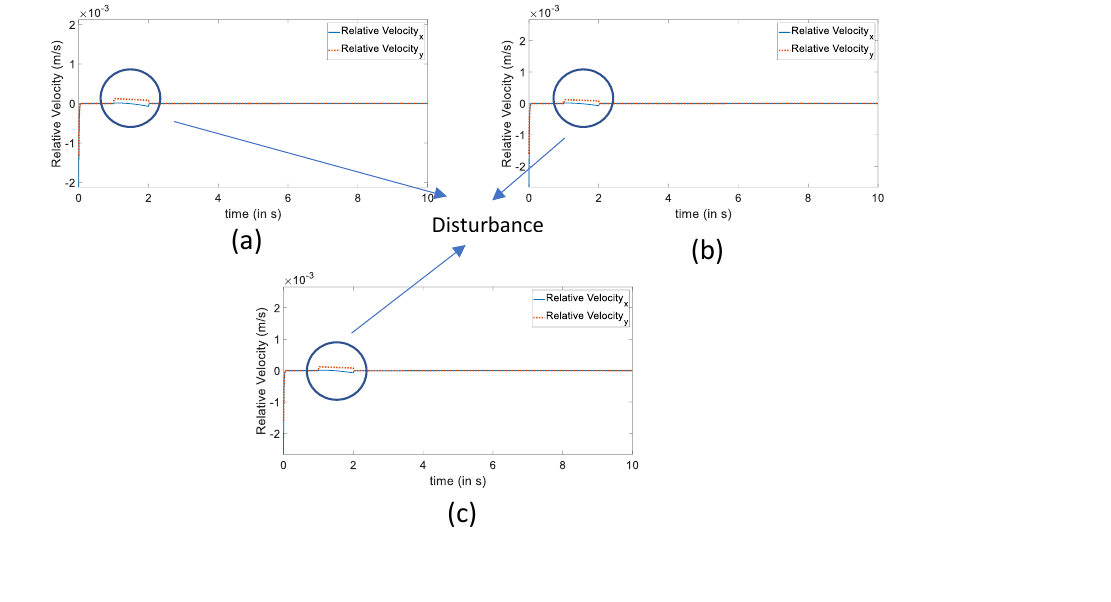}
    \caption{Relative velocity at the finger - object interface for (a) First finger (b) Second Finger (c) Third Finger}
    \label{ellipsoid_another}
\end{figure}
The surface constants for the spherical body in terms of the chart variables is given by:
\begin{equation}
\begin{split}
\label{symbols}
\Gamma_{11k}^1 = 0, 
\Gamma_{12k}^1 =  0,  
\Gamma_{21k}^1 =  0,  
\Gamma_{22k}^1 =  -sin(u_k)cos(u_k) \\
\Gamma_{11k}^2 = 0,  
\Gamma_{12k}^2 =  cot(u_k),  
\Gamma_{21k}^2 =  cot(u_k),  
\Gamma_{22k}^2 =  0 
\end{split}
\end{equation}
where the index `k' represents the $k^{th}$ sphere concerned. \added{The contact curves on the \textit{finger spheres} and the \textit{object sphere} surfaces are described by equation \ref{timegeodesic}. The relative motion between the contacting frames is computed using the Montana's equations of general contact \cite{montana1988kinematics}. This is a kinematic evaluation where persistence of contact throughout the time interval is presumed}. The Montana's equations of general contact relate the relative velocity of contact to the first order derivatives of the contact curves. The first order solution (first order derivative of the contact curve) to equation \ref{contrac} is used to compute the relative velocity at contact. The radius of the object ($r_o$) is 0.1 m whereas the radius of the fingertip ($r_{f_i}$) is 0.04m. The path parameter $\sigma$ is assumed as $0.4t^2+0.001$. A small intercept in $\sigma$ is needed to ensure that the term $\frac{\dot{\sigma}}{\sigma}$ remains finite. The fingertips contact the object at the local object coordinates as $(u_{oi},v_{oi}) = (\frac{\pi}{6},\frac{\pi}{6}),~(\frac{2\pi}{3},\frac{\pi}{6}),~(\frac{\pi}{10},\pi)$. Figure \ref{basic} shows the response of the relative velocity. The initial disturbance is rejected by the system and the relative velocity goes to zero for the three fingers. The modified geodesic equations do not require the initial relative velocity to be zero. As stated earlier, once the relative velocities converge to zero, the fingers necessarily roll on the object until any external disturbance occurs. External disturbances, if induced, can be rejected by selecting an appropriate $\eta$, as in equation \ref{contracaccel}-\ref{contrac}. For this example, we choose, $\eta = 100$. This ensures disturbance to be rejected within 0.04s. \added{Being a kinematic evaluation, it is assumed that the fingertips in contact can impart the desired fingertip acceleration.} Another example case with spherical object and spherical fingers is shown in Figure \ref{contract}. In this case, we disturb the system for 0.1 seconds. During the disturbance, variation of one of the coordinate on each of the fingertip are varied as:
\begin{equation}
\begin{split}
    \ddot{u}_{f_1} = 0.1~rad/s^2\\
    \ddot{u}_{f_2} = 0.1~rad/s^2\\
    \ddot{u}_{f_3} = 0.1~rad/s^2
\end{split}
\end{equation}
The consequent variation of the relative velocity is shown in Figure \ref{contract}. As shown in the figure, the contact curve trajectory rejects the induced disturbance. In this case as well a value of $\eta = 100$ ensures disturbance rejection within 0.04 s after the disturbance ends. For a quicker rejection, a larger $\eta$ can be considered. 
\subsection{Numerical Example with an Ellipsoid}
An ellipsoidal object is used to demonstrate a surface with two non-identical and variable curvatures. The localised chart function for the ellipsoid is given in equation \ref{ellipsoid}. The fingers continue to be spherical in shape, as in the previous example.
\begin{equation}
\label{ellipsoid}
    \mathbf{f_o}(u_{o},v_{o}) = \left[\begin{array}{c} \frac{r_{o1}cos(u_{o})\sqrt{r_{o1}^2-r_{o2}^2sin(v_{o})^2-r_{o3}^2cos(v_{o})^2}}{\sqrt{r_{o1}^2-r_{o3}^2}}\\ r_{o2}sin(u_{o})cos(v_{o}) \\ \frac{r_{o3}sin(v_{o})\sqrt{r_{o1}^2sin(u_{o})^2)+r_{o2}^2cos(u_{oi})^2-r_{o3}^2}}{\sqrt{r_{o1}^2-r_{o3}^2}}\end{array}\right]
\end{equation}
where $r_{oi}$ is the $i^{th}$ radius of the ellipsoid. The Christoffel symbols are calculated from the chart function. The three fingers were placed at the local coordinates of ${(u_{oi}, v_{oi})} = (\frac{2\pi}{3}, \frac{\pi}{2}), (-2\frac{\pi}{3}, -\frac{\pi}{2}), (\frac{\pi}{6}, \frac{\pi}{2})$ rad. In this example, the radii of the ellipsoid are defined as: $r_{o1}~=~0.3~m$, $r_{o2}~=~0.2~m$, $r_{o3}~=~0.1~m$. The path parameter $\sigma$ is selected as $0.4t^2+0.001$.
A disturbance was forced between t = 0.1 and 0.15 seconds. During the disturbance, the trajectory on the fingertip varies as:
\begin{equation}
    \begin{split}
        \ddot{u}_{f_1} = -0.1~rad/s^2,~\ddot{v}_{f_1} = -0.1~rad/s^2 \\
        \ddot{u}_{f_2} = -0.1~rad/s^2,~\ddot{v}_{f_2} = -0.1~rad/s^2 \\
        \ddot{u}_{f_3} = -0.1~rad/s^2,~\ddot{v}_{f_3} = -0.1~rad/s^2
    \end{split}
\end{equation}
The contact coordinate on the objects map the geodesic curve during the disturbance as well. In this example as well, we select $\eta~=~100$. The system rejects the disturbance within 0.015 seconds after the disturbance is over. 
The plot of the relative velocity is shown in Figure \ref{ellipsoid_result}. The disturbance is rejected by the method proposed here.  As observed in Figure \ref{ellipsoid_result}, the relative velocity generated during disturbance is rejected to zero. Consider another case of manipulation, with a different $\sigma$ term, resulting in a slower rate of growth of the contact curve. The $\sigma$ term is given by: $-0.02t^2+0.2t+0.1$. The disturbance between t = 1s and t = 2s is rejected after end of disturbance (Figure \ref{ellipsoid_another}). The disturbance was specified as:
\begin{equation}
    \begin{split}
        \ddot{u}_{f_1} = -0.3~rad/s^2,~\ddot{v}_{f_1} = -0.3~rad/s^2 \\
        \ddot{u}_{f_2} = -0.3~rad/s^2,~\ddot{v}_{f_2} = -0.3~rad/s^2 \\
        \ddot{u}_{f_3} = -0.3~rad/s^2,~\ddot{v}_{f_3} = -0.3~rad/s^2
    \end{split}
\end{equation}

\begin{figure}
    \centering
    \includegraphics[scale=0.5, trim = {2cm, 12cm, 3cm, 5cm}]{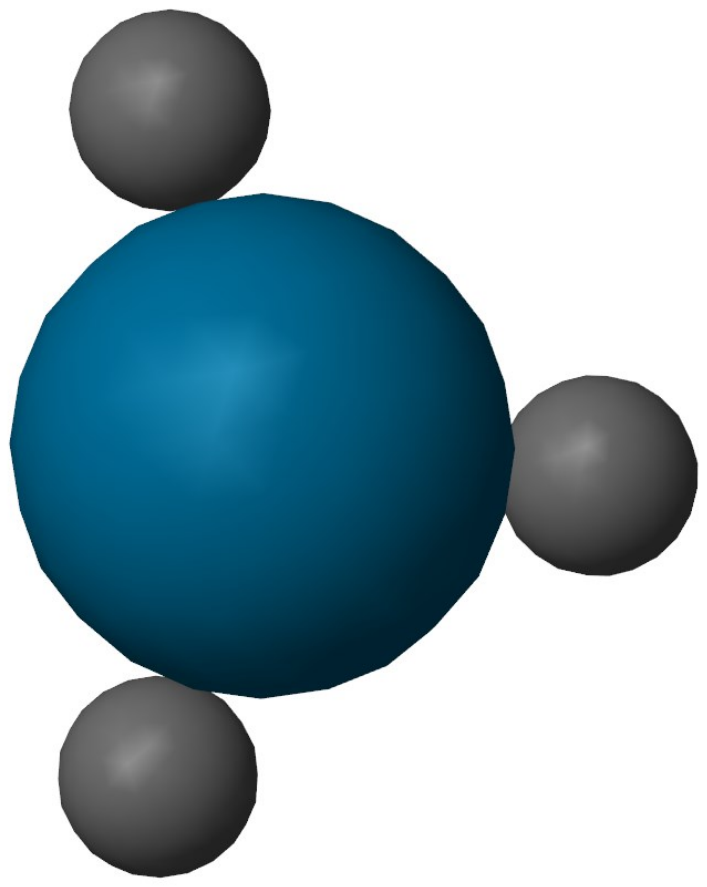}
    \caption{Snip of the dynamic simulation developed using MATLAB Simscape Multibody Physics Engine}
    \label{simul_environment}
\end{figure}

\begin{figure}
    \centering
    \begin{subfigure}[b]{0.48\textwidth}
    \includegraphics[width = 1\textwidth, trim={2cm 19cm 3cm 2cm}, clip]{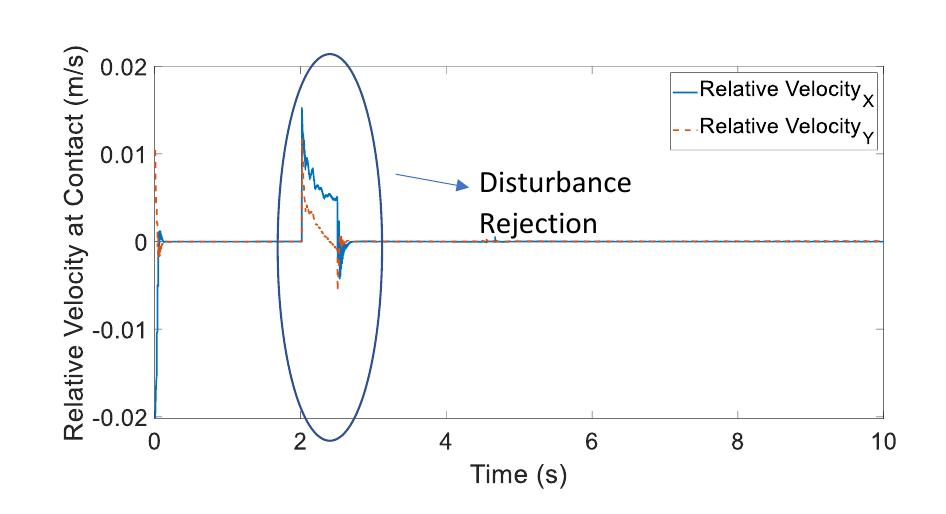}
    \caption{}
    \end{subfigure}\hfill\begin{subfigure}[b]{0.48\textwidth}
    \includegraphics[width=1\textwidth, trim={2cm 18cm 3cm 2cm}, clip]{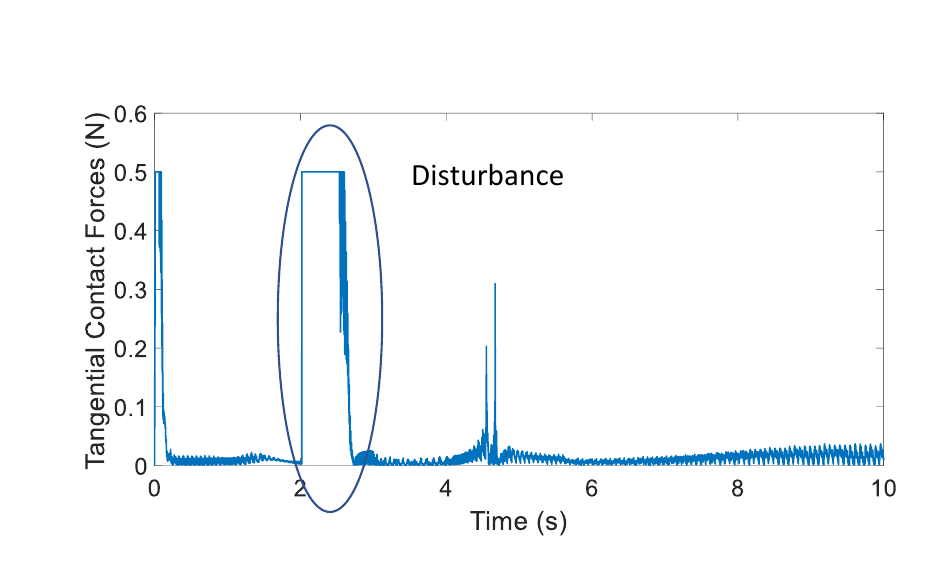}
    \caption{}
    \end{subfigure}
    \begin{subfigure}[b]{0.5\textwidth}
    \includegraphics[width=1\textwidth, trim={2cm 19cm 3cm 2cm}, clip]{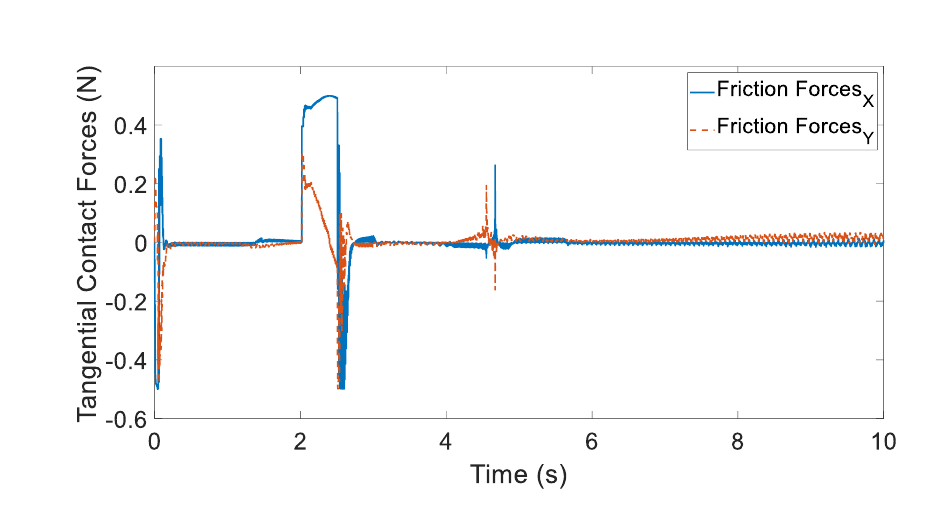}
    \caption{}
    \end{subfigure}
    \caption{(a) Relative Velocity at contact for the first fingertip. The algorithm rejects the disturbance (b)  Magnitude of the tangential forces at the contact. The forces rise when the contact curves are disturbed from geodesic forms. The tangential force is limited to 0.5 N as the normal force at contact is 1 N. (c) Variation of the lateral contact forces along local x and y axes}
    \label{dyn_caseI}

\end{figure}
\begin{figure}
     \centering
     \begin{subfigure}[b]{0.48\textwidth}
         \centering
         \includegraphics[width=1\textwidth, trim={3cm, 18cm, 3cm, 2cm}, clip]{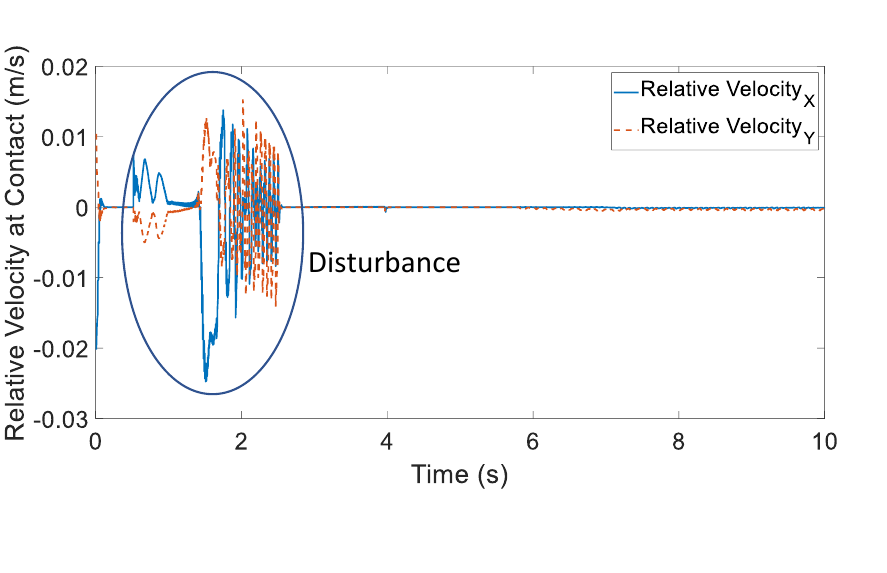}
         \caption{}
     \end{subfigure}\hfill\begin{subfigure}[b]{0.48\textwidth}
         \centering
         \includegraphics[width=1\textwidth,trim={3cm, 18cm, 3cm, 2cm}, clip]{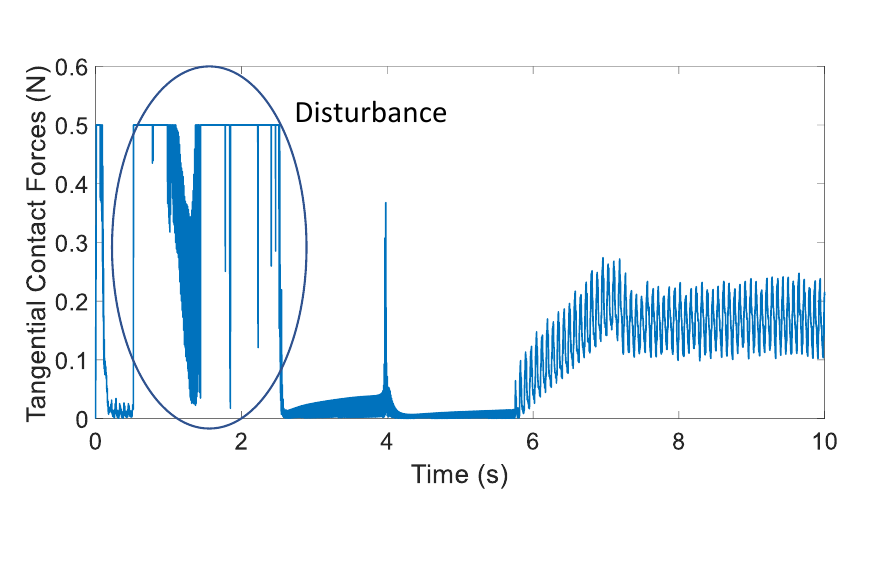}
         \caption{}
     \end{subfigure}
     \caption{(a) Variation of the relative velocity between the contacting frames. The disturbance starts at t = 0.5 s and ends at t = 2.5 s. Post disturbance, the fingertips gain rolling motion. (b) Variation of the norm of the tangential contact force. The tangential forces rise upto 0.5 N (which is 0.5 times the normal force). The normal force magnitude is limited to 1 N }
      \label{dyn_caseIb}
\end{figure}

\begin{figure}
     \centering
     \begin{subfigure}[b]{0.48\textwidth}
         \centering
         \includegraphics[width=1\textwidth, trim={3cm, 19cm, 3cm, 2cm}, clip]{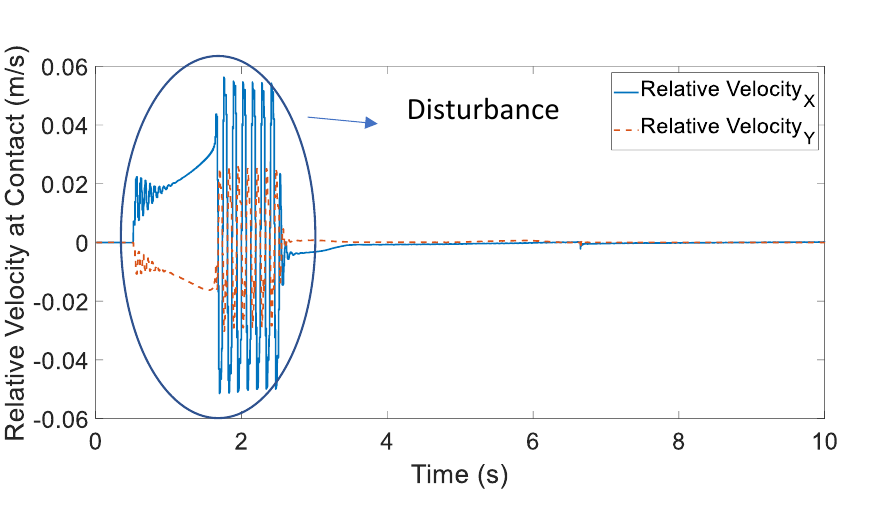}
         \caption{}
     \end{subfigure}\hfill\begin{subfigure}[b]{0.48\textwidth}
         \centering
         \includegraphics[width=1\textwidth,trim={3cm, 19cm, 3cm, 2cm}, clip]{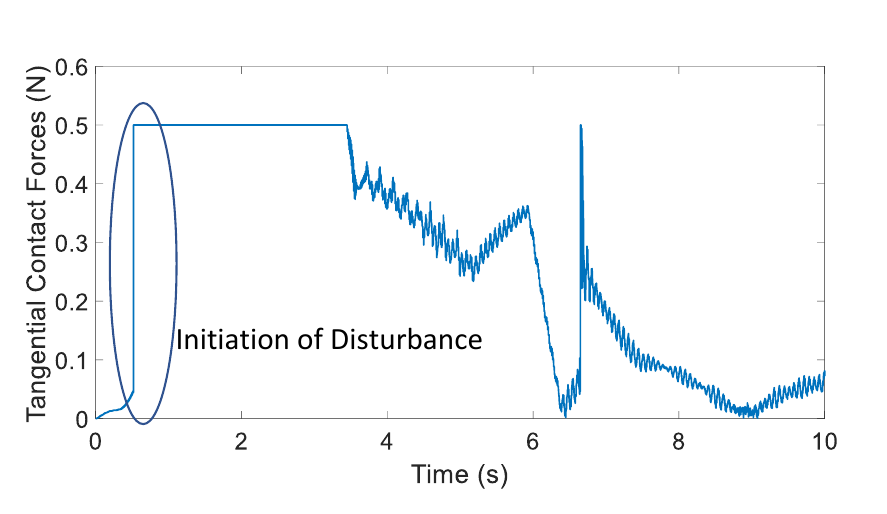}
         \caption{}
     \end{subfigure}
     \caption{(a) Variation of the relative velocity between the contacting frames. The disturbance starts at t = 0.5 s and ends at t = 2.5 s. Post disturbance, the fingertips gain rolling motion. (b) Variation of the norm of the tangential contact force. The tangential forces rise upto 0.5 N (which is 0.5 times the normal force). }
      \label{dyn_caseII}
\end{figure}

\subsection{Dynamic Simulation Examples}
\added{
The derivations presented earlier prove that the contact point traces on the two contacting bodies being geodesics can be used to sustain rolling contact. A modification to the governing equations ensure that the rolling contact can be regained even after a disturbance. Dynamic simulations, which include the contact physics are presented below  and they illustrate the utility of the methodology when the lateral contact forces are limited. }
\par \added{Consider three examples of dynamic simulation, where the permissible normal forces between the surfaces are limited. The MATLAB Simscape Simulation environment \cite{miller2017simscape} was used to simulate the contact physics. The fingertips have been modelled with a spherical radius of 0.04 m, mass 0.261 kg; and the object as a sphere of radius of 0.1 m and mass of 4.18 Kg. The three fingertips initially contact the object at the contact coordinates of $(u_{oi}, v_{oi}) = (0.3886,2.6556),(1.57,2.0944),(-1.57,2.44)$ rad. To accentuate the effect of the limit of the normal force on the rolling contact, the gravitational force is assumed to be zero.  
The initial configuration of the three spherical fingertips contacting the object in the simulation environment is shown in Figure \ref{simul_environment}. The contact interaction is defined through a stiffness of 10,000 N/m, contact damping of 1000 Ns/m and coefficient of friction of 0.5. The normal forces are directed to enforce contact between the object and the fingertip. The fingertips apply a normal force of 1 N to maintain contact.  Correspondingly the norm of the tangential force vector at the $i^{th}$ contact interface ($f_{ti}$) is limited to $\mu f_{Ni}$, where $f_{Ni}$ is the normal force at the $i^{th}$ contact interface. The Montana's equations of general contact motion \cite{montana1988kinematics} are used to relate the motion of the fingertip on the surface of the object to the first derivative of the desired contact curves. The motion of the fingertip is an input to the dynamic simulation. In order to simulate random disturbances, the derivative of the contact curves on the surface of the object are modified during the period of disturbance. The derivative terms are modified as: $\dot{u}_{oi} = \dot{u}_{oi}+\dot{u}_{oi_{dist}}$ and $\dot{v}_{oi} = \dot{v}_{oi}+\dot{v}_{oi_{dist}}$, where $\dot{u}_{oi_{dist}}$ and $\dot{v}_{oi_{dist}}$ modify the actual solutions to equation \ref{contrac} during the period of disturbance.  }
\par \added{For the first case, the fingertips maintain a $\sigma$ profile of $-0.02t^2+0.2t+1$ m/s and the disturbance rejection algorithm presented earlier (equation \ref{contrac}) is implemented. A disturbance is initiated between t = 2.0s and t = 2.5 s at the first fingertip. A disturbance term is added to the derivative of the contact curve on the surface of the object. The added disturbance term is given by: $\dot{u}_{oi_{dist}} = 0.6 rad/s$ and $\dot{v}_{oi_{dist}} = 1 rad/s$. The relative velocity between the contacting frames of the first fingertip and the object surface, is shown in Figure \ref{dyn_caseI}(a). The tangential forces, expressed in the contact frames attached to the surface of the object are shown in Figure \ref{dyn_caseI}(b). The disturbance is rejected within 0.2 seconds post end of the disturbance, regaining rolling contact. The variation of the lateral contact forces along the local x and y axes on the object surface is shown in Figure \ref{dyn_caseI}(c). As another example, disturbance is initiated between t = 0.5 s and t = 2.5 s by injecting additional terms ($\dot{u}_{oi_{dist}} = 0.6 rad/s$ and $\dot{v}_{oi_{dist}} = 0.1 rad/s$) are added to the derivative of the contact curve on the surface of the object. The relative velocity of the first fingertip manipulating the object is shown in the Figure \ref{dyn_caseIb}(a). The tangential contacting forces, referred to the contact frames attached to the surface of the object is shown in Figure \ref{dyn_caseIb}(b). The contact forces reject the disturbance within 0.1 seconds after the disturbances end. }
\par \added{As another illustration, consider a  case of the fingertip rolling on the surface of the object, where the fingertips maintain the $\sigma$ profile of $0.4t^2+0.001$ m/s. The system is disturbed for 2 seconds (from t = 0.5 s to t = 2.5 s). In order to realise the disturbance, additional terms ($\dot{u}_{oi_{dist}} = 0.6 rad/s$ and $\dot{v}_{oi_{dist}} = 0.1 rad/s$) are added to the derivative of the contact curve on the surface of the object. The relative velocity of contact at the first fingertip interface is shown in Figure \ref{dyn_caseII}(a). The variation of the contact forces is shown in Figure \ref{dyn_caseII}(b). The disturbance is rejected within 1 s of the end of the disturbance. The quicker $\sigma$ profile results in slower rejection of the sliding based disturbances. The saturation of lateral forces are sustained until rolling contact is achieved again.}
\begin{figure}[t]
    \centering
    \includegraphics[scale = 0.8, trim={6cm 28cm 0 2cm},clip]{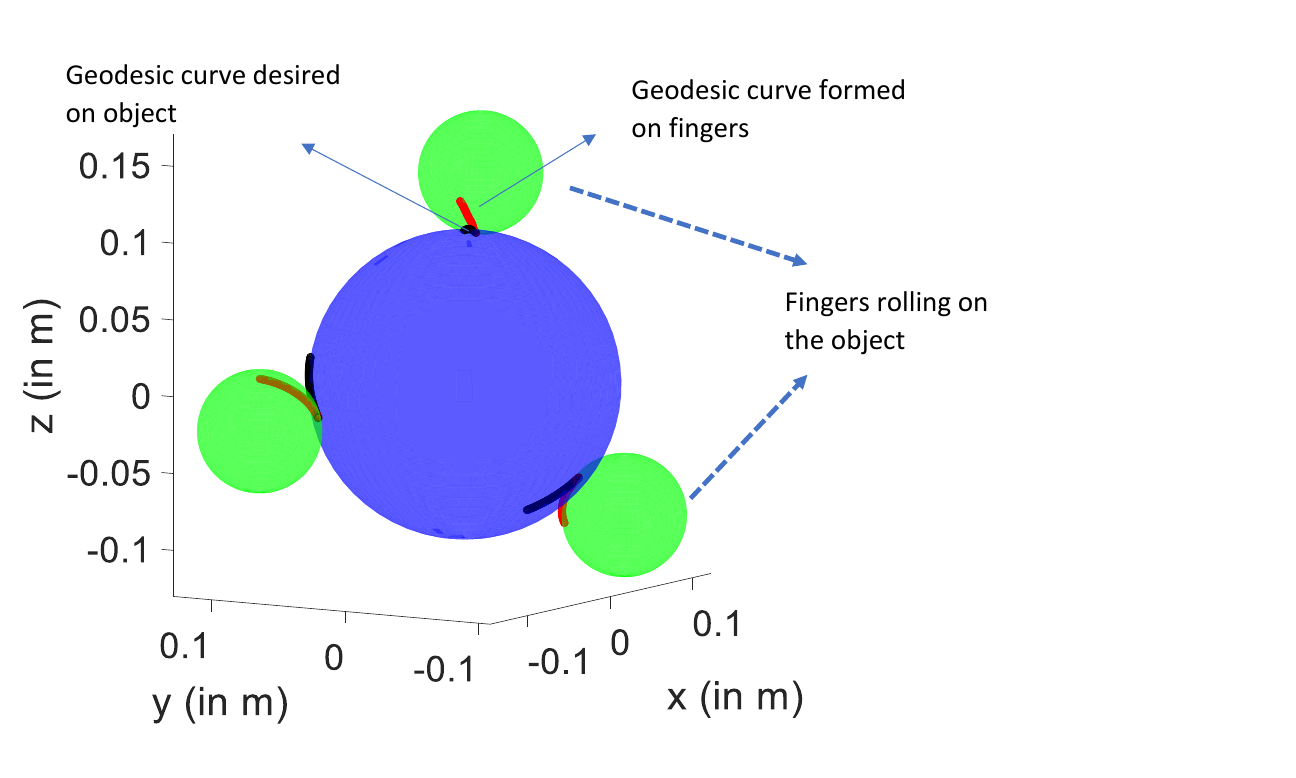}
    \caption{The fingers move on the object with rolling contact. If geodesic curves are traced on the object, then we have geodesic curves on the fingers as well}
    \label{three_rolling_sphere}
\end{figure}
\begin{figure}
    \centering
    \includegraphics[scale = 0.8, trim={2cm 19cm 0 2cm},clip]{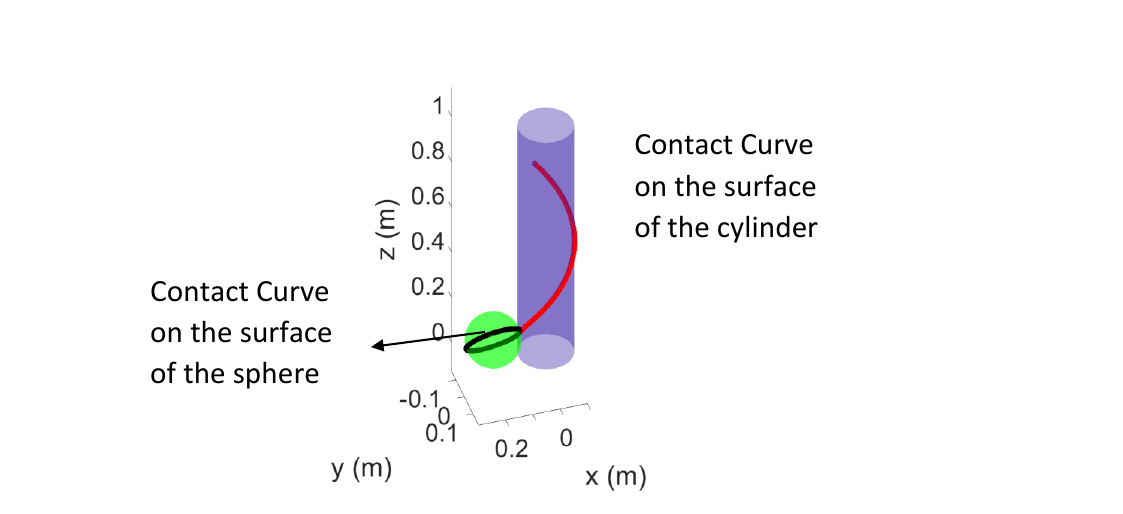}
    \caption{Geodesic helices on the cylinder leads to great circles on the sphere is rolling constraints are known.}
    \label{cyl_sph}
\end{figure}
\section{Corollary}
\label{corollarysec}
The analysis in section \ref{finger} suggests that if the locus of the contact point on both the \added{contacting bodies} follow geodesic curves on respective surfaces, and if slip is zero for a finite time at the start of in-hand manipulation, then slip is zero for all subsequent times. We now prove the converse that if it is known that the contacting bodies have rolling contact, and if the contact point traces a geodesic on body 1 (fingertip), then it results in contact point tracing a geodesic curve on body 2 (object) as well. \par For the case of the contact point tracing a geodesic curve on the surface of body 1 (fingertip), then, equation \ref{final} for the relative acceleration can be written as:
\begin{equation}
\label{corollary}
   \mathbf{a}_{rel} = \left[\begin{array}{ccc}||\frac{\partial \mathbf{f_1}}{\partial u_1}||\frac{\dot\sigma}{\sigma}\dot{u}_1cos\psi-||\frac{\partial \mathbf{f_1}}{\partial v_1}||\frac{\dot\sigma}{\sigma}\dot{v}_1sin\psi-g_1(u_2,\dot{u}_2,v_2,\dot{v}_2) \\ -||\frac{\partial \mathbf{f_1}}{\partial u_1}||\frac{\dot\sigma}{\sigma}\dot{u}_1sin\psi-||\frac{\partial \mathbf{f_1}}{\partial v_1}||\frac{\dot\sigma}{\sigma}\dot{v}_1cos\psi-g_2(u_2,\dot{u}_2,v_2,\dot{v}_2) \\ * 
     \end{array}\right] = \left[\begin{array}{ccc}0 \\ 0 \\ *\end{array}\right]
\end{equation}
After equating first two rows in equation \ref{corollary} , we get:
\begin{equation}
\label{getobjectout}
\begin{split}
g_1(u_2,\dot{u}_2,v_2, \dot{v}_2) = ||\frac{\partial \mathbf{f_1}}{\partial u_1}||\frac{\dot\sigma}{\sigma}\dot{u}_1cos\psi-||\frac{\partial \mathbf{f_1}}{\partial v_1}||\frac{\dot\sigma}{\sigma}\dot{v}_1sin\psi \\
g_2(u_2,\dot{u}_2,v_2,\dot{v}_2) = -||\frac{\partial \mathbf{f_1}}{\partial u_1}||\frac{\dot\sigma}{\sigma}\dot{u}_1cos\psi-||\frac{\partial \mathbf{f_1}}{\partial v_1}||\frac{\dot\sigma}{\sigma}\dot{v}_1sin\psi
\end{split}
\end{equation}
Considering the rolling constraint imposed as $\mathbf{v}_{rel} = \left[\begin{array}{ccc} 0 \\ 0 \\ 0 \end{array}\right]$. The relative velocity kinematics equation \ref{netrelvel} on equating term by term and multiplying by $\frac{\dot{\sigma}}{\sigma}$ yields:
\begin{equation}
\label{velocityeqncor}
\begin{split}
  ||\frac{\partial \mathbf{f_2}}{\partial u_2}|| \frac{\dot{\sigma}}{\sigma} \dot{u}_2 = ||\frac{\partial \mathbf{f_1}}{\partial u_1}||\frac{\dot\sigma}{\sigma}\dot{u}_1cos\psi-||\frac{\partial \mathbf{f_1}}{\partial v_1}||\frac{\dot\sigma}{\sigma}\dot{v}_1sin\psi \\
   ||\frac{\partial \mathbf{f_2}}{\partial v_2}|| \frac{\dot{\sigma}}{\sigma} \dot{v}_2 = -||\frac{\partial \mathbf{f_1}}{\partial u_1}||\frac{\dot\sigma}{\sigma}\dot{u}_1cos\psi-||\frac{\partial \mathbf{f_1}}{\partial v_1}||\frac{\dot\sigma}{\sigma}\dot{v}_1sin\psi
\end{split}
\end{equation}
which can be seen as a solution to $g_1$ and $g_2$. Utilizing the expression of $\mathbf{a}_{rel}$, from equation \ref{finetune} along with equation \ref{corollary}, the portion of acceleration terms in $\mathbf{M_2}\mathbf{a_2}_{E_2}$ reduces to equation \ref{corolla1}
\begin{equation}
\begin{split}
   \left|\left|\frac{\partial \mathbf{f_2}}{\partial u_2}\right|\right|\left(\ddot{u}_2 + {\Gamma^1_{11}}_2\dot{u}_2^2 + 2{\Gamma^1_{12}}_2\dot{u}_2\dot{v}_2+{\Gamma^1_{22}}_2\dot{v}_2^2\right) = \left|\left|\frac{\partial \mathbf{f_2}}{\partial u_2}\right|\right|\frac{\dot{\sigma}}{{\sigma}}\dot{u}_2 \\
    \left|\left|\frac{\partial \mathbf{f_2}}{\partial v_2}\right|\right|\left(  \ddot{v}_2 + {\Gamma^2_{11}}_2\dot{u}_2^2 + 2{\Gamma^2_{12}}_2\dot{u}_2\dot{v}_2+{\Gamma^2_{22}}_2\dot{v}_2^2\right) = \left|\left|\frac{\partial \mathbf{f_2}}{\partial v_2}\right|\right|\frac{\dot{\sigma}}{{\sigma}}\dot{v}_2
\end{split}
    \label{corolla1}
\end{equation}
Multiplying both sides by $\frac{1}{\sigma^2}$, the equation~\ref{corolla1} can be reduced to:
\begin{equation}
\begin{split}
    \frac{1}{\sigma^2}\ddot{u}_2 - \frac{\dot{\sigma}}{{\sigma^3}}\dot{u}_2 + \frac{1}{\sigma^2}{\Gamma^1_{11}}_2\dot{u}_2^2 + 2\frac{1}{\sigma^2}{\Gamma^1_{12}}_2\dot{u}_2\dot{v}_2+\frac{1}{\sigma^2}{\Gamma^1_{22}}_2\dot{v}_2^2 = 0 \\
    \frac{1}{\sigma^2}\ddot{v}_2 + \frac{1}{\sigma^2}{\Gamma^2_{11}}_2\dot{u}_2^2 - \frac{\dot{\sigma}}{{\sigma^3}}\dot{v}_2 + 2\frac{1}{\sigma^2}{\Gamma^2_{12}}_2\dot{u}_2\dot{v}_2+\frac{1}{\sigma^2}{\Gamma^2_{22}}_2\dot{v}_2^2 = 0
\end{split}
    \label{corolla2}
\end{equation}
The equation \ref{corolla2} is the time parameterized equation of geodesic curves (equation \ref{geodesic}) on the object. So, if two bodies are constrained to roll with respect to one another and if the contact curve on one body is a geodesic, then the contact curve on the other body will also be a geodesic curve. The other solution (trivial) is when the absolute velocity and acceleration of both the bodies in the global reference frame are equal to zero. This suggests that the system is at rest and there is no in-hand manipulation. 
\par As a special case, consider that the contact point moves uniformly in time over the path, that is, $\sigma = \text{constant}$.  Then, the right hand side of equation \ref{getobjectout} equates to 0, resulting in $g_1 = 0$ and $g_2 = 0$, which yields $\ddot{u}_i+ \Gamma_{112}^{1}\dot{u}_i^2+2\Gamma_{122}^1\dot{u}\dot{v}+\Gamma_{222}^{1}\dot{v}^2 = 0$ and $\ddot{v}_i +\Gamma_{112}^2\dot{u}_i^2+2\Gamma_{122}^2\dot{u}_i\dot{v}_i+\Gamma_{222}^2\dot{v}_i^2 = 0$, which is the equation of a geodesic curve, with $\sigma = \text{constant}$. This is also true the other way round, that is, if the finger (body 1) maps geodesic curve on the object (body 2) and is rolling on the object, then it will map geodesic curve on the fingertip (body 1). 
\subsection{Examples}
Consider a spherical object with radius $r_o$ = 0.1 m and the finger be spherical tipped with radius $r_f$ = 0.04m. The local charts ($\mathbf{f}(u,v)$) for a sphere are defined as per section \ref{sphere}. Montana's velocity contact equations (\cite{montana1988kinematics}) for rolling contact (relative linear velocities as zero) are used to evaluate the contact system. 
We assume that the rotational trajectory of the object is given by $\mathbf{\omega}^o = [\dot{p}(t),~\dot{p}(t),~\dot{p}(t)]^T$, where $p(t) = a_5t^5+a_4t^4+a_3t^3+a_2t^2+a_1t+a_0$. To generate a minimum jerk trajectory of the object, we use the methodology presented in \cite{zefran1998generation}, where the start and end velocities and acceleration of the object is zero. Considering that the manipulation needs to be done in unit time, a rotational manipulation of the object, consistent with minimum jerk, is given by $p(t)~=~6t^5-15t^4+10t^3$. During simulation, rolling constraint between the object and the fingertip is externally imposed and a geodesic based contact curve is imposed through the differential equations on the object. The trajectory of the contact points on the finger is shown in Figure \ref{three_rolling_sphere}. The curve made on the finger, as expected is a part of the great circle to a sphere, which is a geodesic, hence, demonstrating the corollary.
\added{ As another illustration, consider a sphere rolling onto geodesic helices of a cylinder. The chart for the cylindrical surface is:}
\begin{equation}
\label{cyl}
    \mathbf{f_o}(u_{oi},v_{oi}) = \left[\begin{array}{c}  r_ocos(u_{oi})\\r_osin(u_{oi})\\v_{oi}\end{array}\right]
\end{equation}
\added{Figure \ref{cyl_sph} shows a kinematic evaluation of the same. As seen from Figure \ref{cyl_sph}, the contact curve on the surface of the sphere are great circles. So, geodesics on the cylindrical surface leads to geodesics on the surface of the sphere, when both are under rolling constraints. }

\section{Conclusion}
This paper initially investigates the nature of contact interaction when the contact curves traced on both surfaces are geodesics. The results are derived in the context of in-hand manipulation of objects and is valid for rigid-rigid contacts in general. It is proved that if geodesic based contact curves are imposed on the contacting bodies, then rolling contact is sustained. A corollary is that if rolling contact is maintained between the contacting bodies, and geodesic based contact curve is synthesized on one of the contacting body, then the contact curve on the other body is also a geodesic. It has then been shown that the local differential form of the geodesic based trajectory can be modified to bring the system into a contraction region. This results in rejection of disturbances which perturb the finger-object (or any other set of contacting bodies) from the geodesic trajectory and hence ensures that the system resumes rolling in finite time after the disturbance. The proofs presented are demonstrated using suitable examples and simulations.
\bibliographystyle{model1-num-names}
\bibliography{references}
\end{document}